\documentclass[runningheads]{llncs}

\usepackage{eccv}

\usepackage{eccvabbrv}
\usepackage{multirow}
\usepackage{graphicx}
\usepackage{booktabs}
\usepackage{colortbl}  
\usepackage{xcolor}
\usepackage{array}   
\usepackage{xltabular}
\usepackage[accsupp]{axessibility}  

\usepackage{wrapfig}

\usepackage{hyperref}

\usepackage{orcidlink}
\newcommand{\la}[1]{{\color{blue}#1}}
\newcommand{\ym}[1]{{\color{red}#1}}
\newcommand{\ul}[1]{\underline{\smash{#1}}}

\definecolor{yellow}{RGB}{255,239,213}
\definecolor{green}{RGB}{50,205,50}
\begin{document}

\title{EgoExo-Fitness: Towards Egocentric and Exocentric Full-Body Action Understanding} 

\titlerunning{EgoExo-Fitness}

\author{Yuan-Ming Li\inst{1,3,\ddagger\dagger}\orcidlink{0009-0001-5914-5605}  \and
Wei-Jin Huang\inst{1,3,\dagger}\orcidlink{0009-0008-0600-2206} \and
An-Lan Wang\inst{1,3,\dagger}\orcidlink{0009-0002-5449-6438} \and \\
Ling-An Zeng\inst{3,4}\orcidlink{0000-0002-8125-8024} \and
Jing-Ke Meng\inst{1,3,\ast}\orcidlink{0000-0001-5437-3070}  \and
Wei-Shi Zheng\inst{1,2,3,\ast}\orcidlink{0000-0001-8327-0003}
}

\authorrunning{Y. Li et al.}
\institute{
$^{1}$ School of Computer Science and Engineering, Sun Yat-sen University, China; \\
$^{2}$ Peng Cheng Laboratory, Shenzhen, China; 
$^{3}$ Key Laboratory of Machine Intelligence and Advanced Computing, Ministry of Education, China;\\
$^{4}$ School of Artificial Intelligence, Sun Yat-sen University, China\\
}

\maketitle

{\let\thefootnote\relax\footnotetext{
\scriptsize 
{$\ddagger$}: Project lead. {$\dagger$}: Equal key contributions. {$\ast$}: Corresponding authors. \\ 
Emails: \{liym266, wanganlan\}@mail2.sysu.edu.cn; mengjke@gmail.com; wszheng@ieee.org. 
}}

\begin{abstract}
We present EgoExo-Fitness, a new full-body action understanding dataset, featuring fitness sequence videos recorded from synchronized egocentric and fixed exocentric (third-person) cameras. Compared with existing full-body action understanding datasets, EgoExo-Fitness not only contains videos from first-person perspectives, but also provides rich annotations. Specifically, two-level temporal boundaries are provided to localize single action videos along with sub-steps of each action. More importantly, EgoExo-Fitness introduces innovative annotations for interpretable action judgement--including technical keypoint verification, natural language comments on action execution, and action quality scores. Combining all of these, EgoExo-Fitness provides new resources to study egocentric and exocentric full-body action understanding across dimensions of ``what'', ``when'', and ``how well''. To facilitate research on egocentric and exocentric full-body action understanding, we construct benchmarks on a suite of tasks (\ie, action classification, action localization, cross-view sequence verification, cross-view skill determination, and a newly proposed task of guidance-based execution verification), together with detailed analysis. Data and code are available at \url{https://github.com/iSEE-Laboratory/EgoExo-Fitness/tree/main}.

\end{abstract}

\section{Introduction}
\label{sec:intro}
\vspace{-0.1cm}
\emph{Imagine that one day you put on your smart eyewear and perform fitness activities. Virtual coach embedded in the eyewear can provide feedback on \ul{what}, \ul{when}, and \ul{how well} you performed the action.}
Such a vision draws a scenario in the next generation of AI-assisted fitness exercise, which requires the AI agent to have the ability of egocentric full-body action understanding (EgoFBAU).

However, existing full-body action datasets \cite{finegym,flag3d,humman_eccv22,finediving,aqa7,skating,mtl-aqa,zeng2020hybrid} are predominantly collected from exocentric (third-person) cameras. 
The dependency of fixed exocentric cameras limits the technical practicality in a more flexible manner. 
For instance, it is much more convenient to put on an embodied recording device than to spend time locating a fixed camera.
Inspired by the emerged community of egocentric vision \cite{outlook_ego}, we ask, {{can we embed the virtual coach on your smart eyewear?}} More generally, \textbf{how can we achieve egocentric full-body action understanding?
}

By looking at the field of egocentric video understanding, we find that egocentric full-body action understanding is yet to be well explored due to the lack of datasets. 
Existing egocentric video datasets primarily focus on interactive actions like desktop works \cite{h2o,epic-kitchen, epic-kitchen-100, actionsense,assembly101,ata,ego_procel,holoassist,egoexolearn} (\eg, cooking and assembling) and daily interaction \cite{ego4d,adl,dataego,hoi4d,charadesEgo,assistq} (\ie, interacting with daily objects or humans). The other branch of egocentric datasets \cite{egohumans,egobody,mo2cap2,hps} mainly focuses on body pose estimation and reconstruction rather than understanding full-body action from other dimensions (\eg, verifying the consistency of action sequences and assessing the execution of action).

\begin{figure}[t]
    \centering
    \includegraphics[width=1\linewidth]{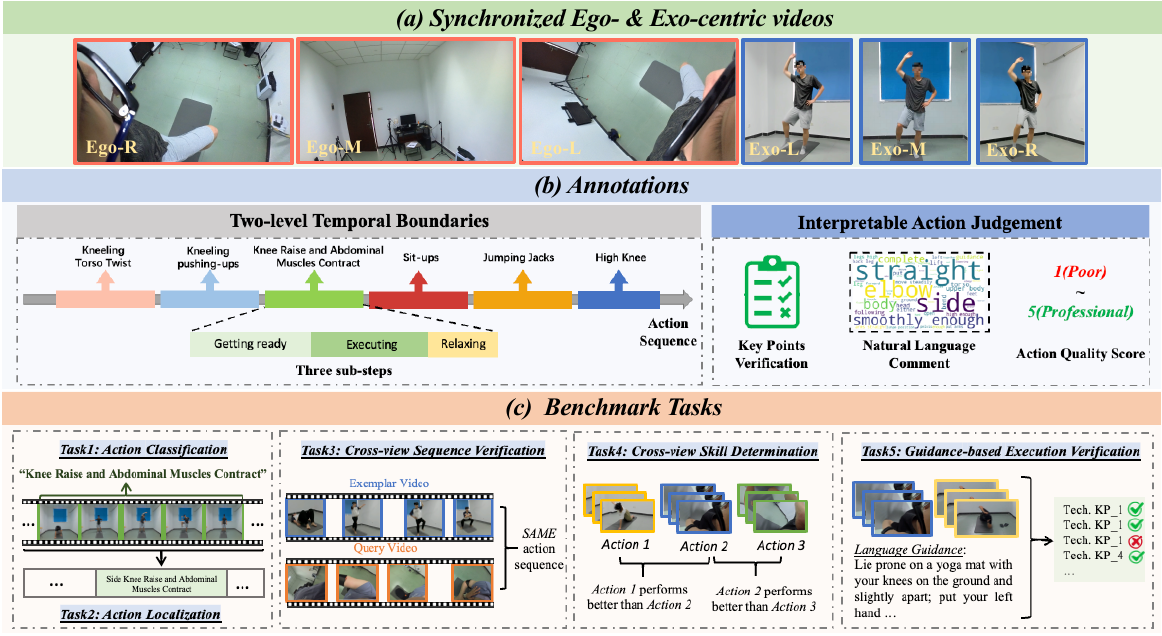}
    \caption{\textbf{An Overview of our work}. (a) We introduce a new video dataset, namely EgoExo-Fitness, which features synchronized egocentric and exocentric videos of fitness activities to support future work on egocentric full-body action understanding. (b) EgoExo-Fitness provides abundant annotations, including two-level temporal boundaries and interpretable action judgement. (c) We benchmark EgoExo-Fitness on five relevant tasks. Zoom in for the best view.}
    \label{fig:dataset_overview}
    \vspace{-0.2cm}
  \end{figure}

To pave the road for future research on full-body action understanding, we focus on fitness activities and present EgoExo-Fitness, a new multi-view full-body action understanding dataset.
An overview of EgoExo-Fitness is shown in \cref{fig:dataset_overview}(a\&b).
The characteristics of EgoExo-Fitness are as follows:

\begin{itemize}
  \item Firstly, on data collection, EgoExo-Fitness features a diverse range of fitness sequence videos recorded by synchronized egocentric and exocentric cameras with various directions;
  \item Secondly, it provides two-level temporal boundaries to localize a single fitness action as well as sub-steps of each action.
  \item More importantly, EgoExo-Fitness introduces annotations on {interpretable action judgement}, 
  including technical keypoint verification, natural language comment, and quality score for each single action execution. 
  To our knowledge, no previous dataset contains such annotations on action judgement.
\end{itemize}

\noindent{Combining} all of these, EgoExo-Fitness
spans 32 hours with 1276 cross-view action sequence videos featuring more than 6000 single fitness actions. With synchronized ego-exo videos and rich annotations, EgoExo-Fitness provides new resources to study egocentric and exocentric full-body action understanding across dimensions of ``what'', ``when'', and ``how well''.

To facilitate research on the line of ego- and exo-centric full-body action understanding, 
as shown in \cref{fig:dataset_overview}(c),
we conduct benchmarks on a suite of tasks, including: \emph{Action Classification}, \emph{Action Localization}, \emph{Cross-View Sequence Verification}, and \emph{Cross-View Skill Determination}. 
More importantly, to further address interpretable action guiding and action assessment, 
we propose {\emph{Guidance-based Execution Verification}}, which aims to infer whether the execution of an action satisfies technical keypoints.
Extensive experiments not only evaluate the effectiveness of baseline methods on the benchmark tasks but also 
pose several challenges for future research.

In summary, the contributions of our work are as follows: 
1) We present EgoExo-Fitness, a new full-body action understanding dataset featuring fitness sequence videos recorded from synchronized ego- and exo-centric cameras; 
2) We introduce rich annotations on EgoExo-Fitness, including two-level temporal boundaries and novel annotations of interpretable action judgement; 
3) We construct benchmarks on five relevant vision tasks, including the newly introduced Guidance-based Execution Verification and extensive experimental analysis.

We expect our dataset and findings can inspire future work on egocentric and exocentric full-body action understanding.

\vspace{-0.2cm}
\section{Related Works}
\label{related_work}
\vspace{-0.2cm}
\subsection{Revisiting Current Datasets}
\vspace{-0.1cm}
We will first revisit today's available full-body action understanding datasets and egocentric video datasets. After that, we will introduce the differences between EgoExo-Fitness and today's datasets.

\vspace{0.1cm}
\noindent{\textbf{Full-Body Action Understanding datasets}.}
Human body movements contain complex motion patterns and technical skills, presenting a series of challenges for Full-Body Action Understanding (FBAU).
To address these challenges, datasets like NTU-RGB+D\cite{ntu-60,ntu-120}, Human3.6M \cite{human36m}, Diving48 \cite{diving48} and FineGym \cite{finegym} are proposed to enable research on recognizing coarse-and-fine human full-body actions.
Beyond recognition, datasets like Diving48-SV \cite{svip} and RepCount \cite{repcount} are present to address tasks (\eg, Sequence Verification and Repetitive Action Counting) that require stronger temporal modeling ability.
Note that technical full-body action videos (\eg, diving and vaulting) will reflect human skills. Hence, in recent years, datasets for Action Assessment, like AQA-7\cite{aqa7}, FineDiving \cite{finediving}, LOGO\cite{logo}, are introduced to study the subtle skill differences between action videos.
Another branch of datasets \cite{flag3d,aifit,humman_eccv22} focuses on estimating or reconstructing 3D human poses from full-body action videos, achieving the development of Virtual Reality.
Though great progress has been achieved, today's full-body action understanding datasets mainly assume that human full-body action videos are captured by exocentric cameras. Such an assumption limits further exploration in more flexible settings.
Moreover, some datasets (\eg, WEAR \cite{wear} and 1st-basketball \cite{1st-basketball}) propose to understand sports and fitness activities from egocentric viewpoints. However, these datasets are limited by their scales and task-specific annotations.

\vspace{0.1cm}
\noindent{\textbf{Egocentric Video Understanding Datasets}.}
Egocentric Video Understanding has great application values for AR/VR and Robotics.
Most existing datasets focus on interactive actions: 1) tabletop activities in kitchen \cite{epic-kitchen, epic-kitchen-100, actionsense,egtea,cmu-mmac} or on a static working platform \cite{h2o,assembly101,ata,ego_procel,meccano}; 2) daily activities interacting with daily objects \cite{assistq,assistsr,charadesEgo,hoi4d,epic-tent,egoexolearn} or individuals \cite{you2me,ryan2023egocentric,easycom}. 
Although recently proposed Ego4D \cite{ego4d} expands beyond interactive activities to a wider variety of daily activities, works on this branch of datasets rarely focus on egocentric full-body action understanding.
Another branch of work aims to estimate full-body pose from egocentric videos, and several datasets \cite{egoego,egobody,unrealego,egoscene,egohumans}  are released.

Different from existing datasets, EgoExo-Fitness features synchronized egocentric and exocentric videos of full-body fitness actions and provides rich annotations (especially novel annotations of interpretable action judgement) for future research on understanding ego- and exo-centric full-body actions across the dimensions of ``what'', ``when'', and ``how well''.

It is worth noting that a concurrent large dataset, Ego-Exo4D \cite{egoexo4d}, also contains ego-exo full-body (physical) action videos and annotations on \emph{how well an action is performed}. EgoExo-Fitness still has its values: (1) It focuses on a novel scenario (\ie, natural fitness practising); (2) We provides novel annotations (e.g., technical keypoint verification), enabling studies on new tasks (\eg, Guidance-based Execution Verification). We will provide detailed comparisons between Ego-Exo4D and our work in \cref{sec:comparison_datasets} and Appendix {A4.2}.

\vspace{-0.2cm}
\subsection{Revisiting Relevant Tasks}
\label{sec:revisit_task}
\vspace{-0.1cm}
In this part, we will present the relationships between the benchmarks of EgoExo-Fitness and relevant tasks. We will further introduce the motivations and set-ups in detail when introducing the benchmarks.

\vspace{0.1cm}
\noindent{\textbf{Action Classification \& Localization}.}
As the fundamental tasks in video action understanding, action classification \cite{lin2024rethinking,timesformer,i3d,action_recog_survey} and temporal action localization \cite{tadtr,actionformer,tal_survey,du2022weakly} are widely explored in previous work. In our work, we benchmark EgoExo-Fitness on action classification and localization to present the gap and characteristics across views in full-body action understanding.

\vspace{0.1cm}
\noindent{\textbf{Sequence Verification}}.
Sequence Verification (SV) \cite{svip, weakly-svr,he2023collaborative} is proposed to study the action order of sequential videos under a scenario that precise temporal annotations are not provided.
Today's benchmarks on SV rather focus on exocentric-SV (\ie, COIN-SV and Diving48-SV) or egocentric-SV (\ie, CSV).
In this work, we present the first benchmark on cross-view sequence verification and provides extensive experimental analysis.

\vspace{0.1cm}
\noindent{\textbf{Action Assessment}.}
Existing datasets on Action Assessment (or Skill Determination) are mainly based on videos from either ego-cameras \cite{whos_better,jigsaws} or exo-cameras \cite{finediving,mtl-aqa,aqa7,logo,skating,raan,zeng2020hybrid,gao2020asymmetric,zeng2024multimodal}, which leads to single-view assessment ability. Also, today's popular datasets only provide the annotations on action scores or paired rankings, which is unable existing work to directly explore the interpretability of the predicted results. 
To address the first issue, we introduce the first benchmark on Cross-View Skill Determination. For the second issue, we propose a novel task, Guidance-based Execution Verification, which aims to verify whether the execution of an action satisfies the given technical keypoints.

\vspace{-0.2cm}
\section{EgoExo-Fitness Dataset}

\label{sec:dataset}
\vspace{-0.1cm}
In this section, we introduce the EgoExo-Fitness dataset in detail. We will start by describing the recording system in \cref{sec:rec_system}, data collection in \cref{sec:data_collection}, and annotations in \cref{sec:annotations}. After that, we present the statistics and comparisons with related datasets in \cref{sec:statistics} and \cref{sec:comparison_datasets}, respectively.

\begin{figure}[t]
  \centering
  \includegraphics[width=0.9\linewidth]{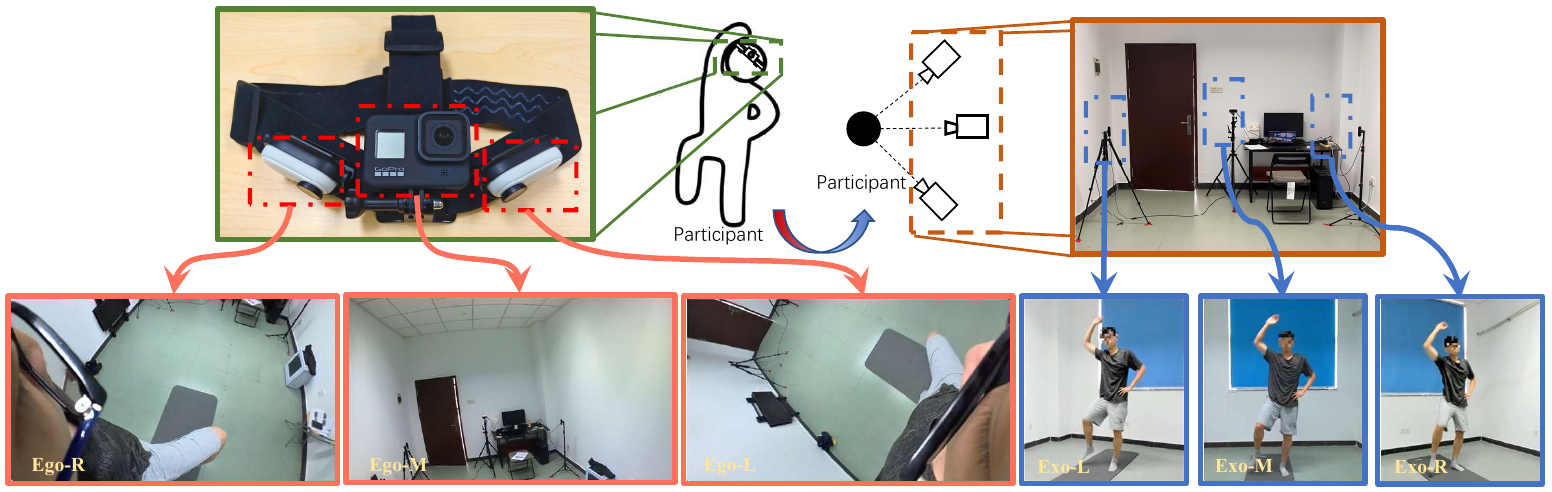}
  \vspace{-0.2cm}
  \caption{\textbf{The setup of our recording system.} We capture forward and downward egocentric videos by developing a headset containing three action cameras. To record exocentric videos, three cameras are located at the front, left-front and right-front sides of the actor. Zoom in for the best view.}
  \label{fig:recording_system}
  \vspace{-0.3cm}
\end{figure}

\vspace{-0.2cm}
\subsection{Recording System}
\label{sec:rec_system}
\vspace{-0.1cm}
We build a recording system for EgoExo-Fitness to capture action videos from egocentric and exocentric views. \cref{fig:recording_system} shows the setups of our recording system.
For egocentric video capturing, we design a headset with multiple cameras to record videos from forward and downwards views.
Specifically, we use a GoPro to record the forward (\ie, Ego-M) view of participants and apply two Insta-Go3 cameras to record the left-downward (\ie Ego-L) and right-downward (\ie, Ego-R) views. For the exocentric cameras, we locate them at the participants' front (\ie, Exo-M), left-front (\ie, Exo-L), and right-front (\ie, Exo-R) sides and ensure they can record full-body actions completely.
All cameras are synchronized manually using a timed event that is visible from them.

\vspace{-0.2cm}
\subsection{Action Sequence \& Recording Protocols}
\label{sec:data_collection}
\vspace{-0.1cm}
Following FLAG3D \cite{flag3d} and HuMMan \cite{humman_eccv22}, we select 12 types of fitness actions based on various driving muscles (\ie, \emph{chest},  \emph{abdomen},  \emph{waist},  \emph{hip}, and  \emph{whole body}). All selected actions are listed in \cref{tab:motion_category}. In the remaining part of the paper, for convenience, we will use the abbreviation to present the actions.
Furthermore, to enrich the temporal diversity of the recorded videos, we define 86 action sequences by randomly combining 3 to 6 different actions. For example, ``starting with \emph{Push-ups}, then \emph{Sit-ups}, finally \emph{High Knee}'' is an action sequence with three fitness actions. 
For details of the action sequences, please refer to Appendix A2.

\vspace{0.1cm}
\noindent{\textbf{Recording Protocols}.}
Before recording, action sequences will be randomly allocated to the participants. Since we are interested in capturing the natural actions of the participants, we only provide the text guidance in advance.
During recording, the participants are asked to put on the headset and continuously complete all actions in the allocated action sequence. For each action, the participants are required to repeat it at least 4 times.

\begin{table}[t]
    \begin{center}
    \caption{\textbf{Recorded fitness actions.} Abbr.: the abbreviation of the fitness action.} 
    \label{tab:motion_category}
    \vspace{-0.2cm}
    \scalebox{0.7}{
    \begin{tabular}{lr||lr||lr}
    \toprule
    \textbf{Action types}                                & \textbf{Abbr.}  & \textbf{Action types}                                & \textbf{Abbr.} & \textbf{Action types}                                & \textbf{Abbr.}\\
    \midrule
    \,\,1:\,Kneeling Push-ups                              & KPU            & \,\,5:\,Shoulder Bridge                                & SB & \,\,10:\,Jumping Jacks                                & JJ\\
    \,\,2:\,Push-ups                                       & PU             & \,\,6:\,Sit-ups                                        & SU & \,\,11:\,High Knee                                    & HK\\
    \,\,3:\,Kneeling Torso Twist                           & KTT            & \,\,7:\,Leg Reverse Lunge                             & LRL & \,\,12:\,Clap Jacks                                   & CJ\\
    \,\,4:\,Knee Raise and Abdomi-\,\,\,\,\,\,\,\,\,\,     & KRAMC          & \,\,8:\,Leg Lunge with Knee Lift\,\,\,\,\,\,\,\,\,\,  & LLKL\\
    \,\,\,\,\,\,\,nal Muscles Contract                     &                & \,\,9:\,Sumo Squat                                    & SS\\
    \bottomrule
    \end{tabular}
    }
    \end{center}
    \vspace{-0.3cm}
\end{table}
\vspace{-0.2cm}
\subsection{Annotations}
\label{sec:annotations}
\vspace{-0.1cm}
To support future work on EgoFBAU, EgoExo-Fitness provides annotations for two-level temporal boundaries and interpretable action judgement.

\begin{figure}[t]
    \centering
    \includegraphics[width=\linewidth]{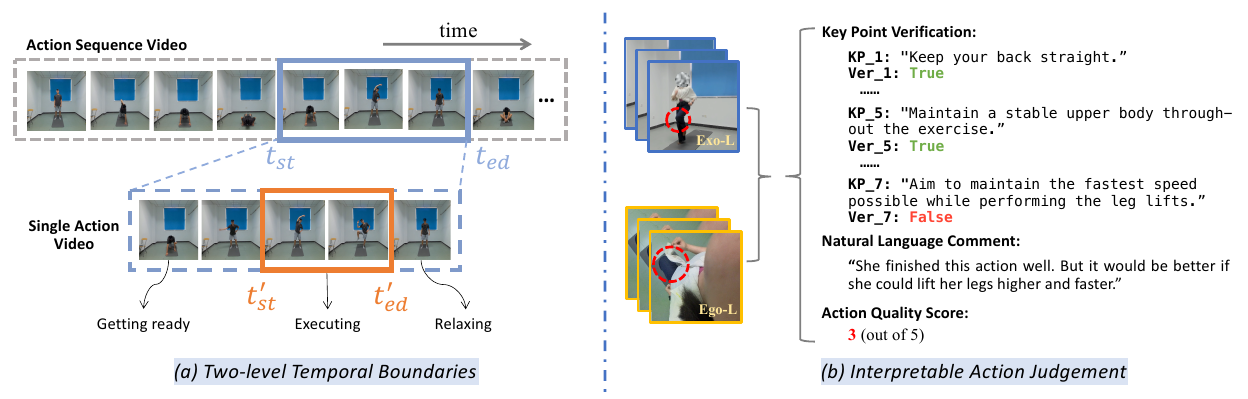}
    \caption{\textbf{Overview of annotations setups.} 
    \textbf{(a)} Two-level temporal boundaries are provided. Specifically, $1^{st}$-level boundaries ($t_{st}$ and $t_{ed}$) localize the single actions from the action sequence video (obtaining single action videos). After that, $2^{nd}$-level boundaries ($t^{'}_{st}$ and $t^{'}_{ed}$) separate every single action into three sub-steps (\,  i.e., getting ready, executing, and relaxing).
    \textbf{(b)} EgoExo-Fitness contains three types of annotations on action judgement, including keypoint verification (KP: keypoint; Ver: verification result), natural language comments, and action quality scores.
    Zoom in for the best view.}
    \label{fig:annotation_level}
    \vspace{-0.3cm}
\end{figure}

\vspace{0.1cm}
\noindent{\textbf{Two-level Temporal Boundaries.}}
To enable studies on action boundaries and action orders, we adopt a two-stage strategy to collect the annotations for the two-level temporal structures of each instance.
To begin with, given an action sequence video (containing 3 to 6 continuous actions) from any camera view, annotators are asked to accurately locate the start and end time (\eg, $t_{st}$ and $t_{ed}$ in \cref{fig:annotation_level}(a)) of each complete action so that a single action video can be obtained.
After that, for each single action video, the annotators are asked to separate the video into three steps(\ie, \emph{Getting ready}, \emph{Executing} and \emph{Relaxing}) and annotate the start and end time (\eg, $t^{'}_{st}$ and $t^{'}_{ed}$ in \cref{fig:annotation_level}(a)) of the \emph{Executing} steps.

\vspace{0.1cm}
\noindent{\textbf{Interpretable Action Judgement.}}
Our motivation for providing this series of annotations is two-fold. 
First, it is easy for human experts to compare an action video and the text guidance to conclude whether the actor followed the guidance or not and point out which technical keypoint in the text guidance is missed during the execution. However, such ability has rarely been studied in existing video-language answering and video-language retrieval works.
Second, although Action Assessment \cite{aqa7,skating,mtl-aqa,finediving,whos_better,raan,con_aqa} has been studied for many years, and great progress has been achieved, interpretable action assessment based on language annotations has never been explored due to the lack of well-collected dataset. Also, existing action assessment work is limited to ego-only or exo-only scenarios due to the collecting manner of the datasets.

To address these issues, we develop a web-based annotation tool for EgoExo-Fitness, and collect three categories of interpretable action judgement annotations (\ie, \emph{Technical keypoints Verification}, \emph{Natural Language Comment}, and \emph{Action Quality Score}) step by step. We will introduce the details as follows.

\textbf{(1) \emph{Technical Keypoint Verification.}} A paragraph of text guidance on fitness action can be divided into several technical keypoints. By following the keypoints, one can achieve the goal of exercise while avoiding physical injury. In EgoExo-Fitness, we provide the verification annotations on the keypoints for located single actions in the following three steps. First, following FLAG3D \cite{flag3d}, we provide a paragraph of text guidance for recorded actions. Second, we prompt LLM (\eg, ChatGPT) to separate the guidance into several keypoints. Third, we ask the annotators to verify an action by comparing the execution with the technical keypoints. Given an action and a technical keypoint, if the action satisfies the keypoint, an annotation of ``\textcolor{green}{True}'' will be noted (otherwise ``\textcolor{red}{False}'').

\textbf{(2) \emph{Natural Language Comment.}} After verifying the technical keypoints, the annotators are asked to write a paragraph of natural language comment on how well the participant finished the action.  
We require that the comments should reflect the verification results from the previous step. 
Additionally, annotators are asked to write a few sentences on how to improve the movements following their subjective appraisals.

\textbf{(3) \emph{Action Quality Score.}} Finally, the annotators are asked to score the actions from 1 to 5 (worst to best) based on the technical keypoint verifications and comments they have made.

\cref{fig:annotation_level}(b) gives an example of the annotations on interpretable action judgement. As shown in the frames cropped from ego- and exo-centric videos, the participant is executing ``high knee''. Though generally performed well, it can be observed from the video that her legs are not lifted high and fast enough (\ie, \emph{red circles} on the cropped frames). Therefore, relevant keypoints will be verified and annotated as \emph{False} (\ie, KP\_7 in \cref{fig:annotation_level}(b)). Besides, natural language comments on the execution will be provided together with improvement advice (\ie, ``It would be better if she could lift her legs higher and faster''). Finally, a subjective action quality score (\ie, 3) is annotated by the annotator.

To ensure the annotation quality, for each single action video, we employ at least two human experts to provide interpretable action judgement annotations.

\vspace{-0.2cm}
\subsection{Statistics}
\vspace{-0.1cm}
\label{sec:statistics}

\begin{figure}[t]
  \centering
  \includegraphics[width=1\linewidth]{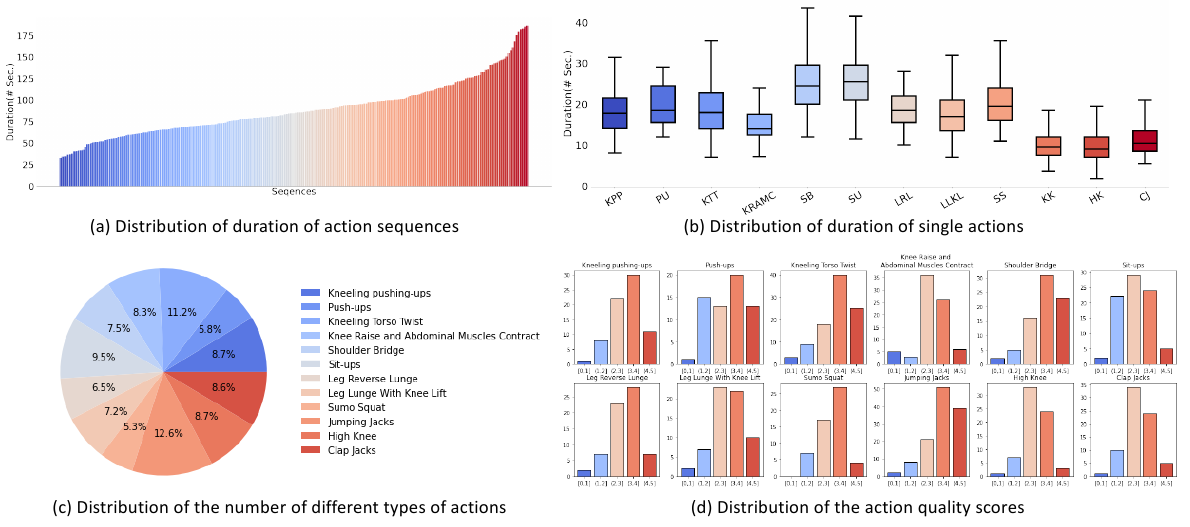}
  \caption{Statistics of the proposed EgoExo-Fitness dataset. }
  \label{fig:statistics}
  \vspace{-0.3cm}
\end{figure}

\noindent{\textbf{Number of recordings and Duration.}}
EgoExo-Fitness collects 1276 cross-view action sequence videos from 86 action sequences, which spans about 32 hours. With two-level temporal boundaries, 6131 single actions are located, 
\cref{fig:statistics}(a \& b) present the duration distribution of action sequence videos and single action videos. 
The duration of action sequence videos is widely distributed between 33 and 186 seconds, and most actions last from 10 to 30 seconds.
The distribution of the number of different types of action is shown in \cref{fig:statistics}(c), where ``Jumping jacks'' takes up the highest proportion of takes (\ie, 12.6\%), and the action takes up the fewest proportion of takes is ``Sumo Squat'' (\ie, 5.3\%).

\vspace{0.1cm}
\noindent{\textbf{Action Quality Score Distribution.}}
We also analyze the distribution of action quality scores for each type of action in \cref{fig:statistics}(d). Here, the score for each single action is calculated by averaging the scores annotated by all annotators. 

\begin{table}[t]
    \caption{\textbf{Comparison with related datasets}. We compare existing datasets on scenarios, annotations and durations. For fair comparison, we select a subset of Ego4D with scenarios of technical full-body action (\eg, dancing, climbing, working-out). SI: Social Interaction. $\divideontimes$: MTL-AQA and Ego4D contains captions and narrations on \emph{what have happened} rather than \emph{how well an action has been done} as in EgoExo-Fitness. }
    \vspace{-0.2cm}
    \label{tab:comprison_datasets}
    \scriptsize
        \centering
        \scriptsize
        \resizebox{0.9\linewidth}{!}{
        \begin{tabular}{l|c|c|cc|ccccc|c}
        \toprule
        \multirow{2}{*}{\textbf{Datasets}} & \multirow{2}{*}{\textbf{Public.}} & \multirow{2}{*}{\textbf{Scenarios}} & \multirow{2}{*}{\textbf{Ego}}      & \multirow{2}{*}{\textbf{Exo}}          & \multirow{2}{*}{\textbf{Step}} & {\textbf{Text}} & \textbf{Keypoint} & \multirow{2}{*}{\textbf{Comment}} & \multirow{2}{*}{\textbf{Score}} &\multirow{2}{*}{\textbf{Duration}}\\ 

        \multirow{2}{*}{} & \multirow{2}{*}{} & \multirow{2}{*}{}   & \multirow{2}{*}{}      & \multirow{2}{*}{}        & \multirow{2}{*}{} & \textbf{guidance} & \textbf{verification} & \multirow{2}{*}{} & \multirow{2}{*}{} & \multirow{2}{*}{}\\ 
        
        \midrule
        \multicolumn{11}{l}{\emph{\textbf{Exocentric full-body action datasets}}} \\
        MTL-AQA\cite{mtl-aqa} & CVPR20 & Diving & &  \checkmark  & &     &&$\divideontimes$&\checkmark& 1.6h \\
        FineGym\cite{finegym}  & CVPR20 & Sports & & \checkmark  &      \checkmark  & &&&& 161h\\
        FineDiving\cite{finediving}& CVPR22 & Diving & & \checkmark  &    \checkmark & &&& \checkmark & 3h\\
        FLAG3D(virtual)\cite{flag3d}    & CVPR23      & Fitness   &            & \checkmark  &            & \checkmark & &&& $\sim$185h\\ 
        FLAG3D(real)\cite{flag3d}  & CVPR23        & Fitness   &            &  \checkmark &              &            \checkmark & &&&69h\\ 
        \midrule
        \multicolumn{11}{l}{\emph{\textbf{Egocentric full-body action datasets}}} \\
        1st-basketball\cite{1st-basketball} & ICCV17 & Basketball & \checkmark & &&&&& \checkmark & 10.3h \\
        Ego4D(tfba)\cite{ego4d} & CVPR22 & Daily
        & \checkmark &                                   & \checkmark &            &            &  $\divideontimes$   &  & 172h \\
        WEAR \cite{wear} & ArXiv23 & Fitness & \checkmark & & \checkmark&&&&&15h\\
        \rowcolor{yellow} EgoExo-Fitness(Ours)     &     & Fitness   & \checkmark   & \checkmark &  \checkmark & \checkmark & \checkmark & \checkmark & \checkmark & 32h\\
        \bottomrule
        \end{tabular}
        }
\end{table}

\vspace{-0.2cm}
\subsection{Comparison with Related Datasets}
\label{sec:comparison_datasets}
\vspace{-0.1cm}
We compare EgoExo-Fitness with popular ego- and exo-centric full body action understanding datasets \cref{tab:comprison_datasets}.
EgoExo-Fitness is the first dataset that features synchronized exo- and ego-centric videos to address egocentric full-body action understanding across dimensions of ``what'', ``when'' and ``how well''. 
Additionally, EgoExo-Fitness introduces novel annotations on interpretable action judgement (\ie, keypoint verifications and comments on how well an action is performed), which make EgoExo-Fiteness different from existing datasets.
With synchronized videos and rich annotations, EgoExo-Fitness provides new resources for studying view characteristics, cross-view modeling, and action guiding.
It is also notable that for a fair comparison, we select a subset from Ego4D \cite{ego4d}, which includes scenarios of technical full-body actions (e.g., dancing, and working-out).

\begin{table}[t]
  \caption{\textbf{Comparison with the concurrent Ego-Exo4D \cite{egoexo4d} dataset}. The proposed EgoExo-Fitness collects videos of a new scenario and augments data with novel annotations. For fair comparison, scenarios of full-body actions are considered.}
  \vspace{-0.2cm}
  \label{tab:comprison_datasets_egoexo_partial}
  \scriptsize
      \centering
      \scriptsize
      \resizebox{0.8\linewidth}{!}{
      \begin{tabular}{l|c|ccccc|c}
      \toprule
      \multirow{2}{*}{\textbf{Datasets}} & \multirow{2}{*}{\textbf{Scenarios}}     & \multirow{2}{*}{\textbf{Step}} & \textbf{Text} & \textbf{Keypoint} & \multirow{2}{*}{\textbf{Comment}} & \multirow{2}{*}{\textbf{Score}} &\multirow{2}{*}{\textbf{Duration}}\\ 

      \multirow{2}{*}{} & \multirow{2}{*}{}          & \multirow{2}{*}{} & \textbf{guidance} & \textbf{verification} & \multirow{2}{*}{} & \multirow{2}{*}{} & \multirow{2}{*}{}\\ 
      
      \midrule
      \multirow{4}{*}{Ego-Exo4D v2 \cite{egoexo4d}}          & Basketball &  & & & \checkmark & \checkmark & 74h \\
      \multirow{4}{*}{}           & Climbing &  &  & & \checkmark & \checkmark & 93h \\
      \multirow{4}{*}{}           & Soccer &  &  & & \checkmark & \checkmark & 66h\\
      \multirow{4}{*}{}           & Dancing &  &  & & \checkmark & \checkmark & 106h\\
      \midrule
      \cellcolor{yellow}EgoExo-Fitness(Ours)    & \cellcolor{yellow}Fitness    & \cellcolor{yellow}\checkmark & \cellcolor{yellow}\checkmark & \cellcolor{yellow}\checkmark & \cellcolor{yellow}\checkmark & \cellcolor{yellow}\checkmark & \cellcolor{yellow}32h\\
      \bottomrule
      \end{tabular}
      }
      \vspace{-0.3cm}
\end{table}

Note that a recent proposed large-scale dataset Ego-Exo4D \cite{egoexo4d} also contains \emph{full-body (physical)} action videos collected by synchronized ego-exo cameras. Besides, they both attend to \emph{how well} an action is performed and propose novel corresponding annotations. What makes our dataset different from Ego-Exo4D lines in the following ways (also shown in \cref{tab:comprison_datasets_egoexo_partial}): 
{\textbf{(1) \emph{New scenario is focused on}}}. We focus on the scenario of natural fitness practising and collect dynamic action sequence videos (containing 3 to 6 different actions). However, in Ego-Exo4D, a video is only associated with one type of action/task.
This makes our dataset better suited for ego-exo full-body action studies on action boundaries and orders.
{\textbf{(2) \emph{Novel annotations are provided}}}. We provide text guidance and technical keypoints verification (both NOT included in Ego-Exo4D), which offer another branch of intuitive and detailed identifications of \emph{what is done well} and \emph{what can be improved} in executions than expert commentary in Ego-Exo4D. Such annotations enable the pioneering exploration of interpretable action assessment.
{\textbf{(3) \emph{Other unique characteristics}}}. We also provide videos captured from two downward ego-cameras for capturing more body details in movements, and annotations of two-level temporal boundaries to enable benchmark constructions.

For more details about dataset comparisons, please refer to Appendix A4.

\vspace{-0.2cm}
\section{Benchmarks}
\label{sec:benchmarks}
\vspace{-0.1cm}

With synchronized ego-exo videos and rich annotations, EgoExo-Fitness can provide resources for studies of view characteristics, cross-view modeling, and action guiding. 
To benefit future research of these directions on EgoExo-Fitness, we conduct benchmarks on \emph{Action Classification} (\cref{sec:recognition}), \emph{Cross-View Sequence Verification} (\cref{sec:cvsv}), and a newly proposed \textit{Guidance-based Execution Verification} (\cref{sec:gev}). 
EgoExo-Fitness also supports \emph{Action Localization} and \emph{Cross-View Skill Determination}, which are presented in Appendix A3.

\vspace{-0.2cm}
\subsection{Action Classification}
\label{sec:recognition}
\vspace{-0.1cm}

We select Action Classification \cite{action_recog_survey}, the fundamental task of video action understanding, to study view gap and view characteristics on EgoExo-Fitness.

\vspace{0.1cm}
\noindent\textbf{Task Setups.}
We share the same task setups with previous works on action classification, \ie, to predict the type of fitness action given a trimmed single action video from either ego-or-exo viewpoint.

\vspace{0.1cm}
\noindent{\textbf{Baseline Models.}}
We apply three baseline models in Action Classification benchmark: (1) I3D \cite{i3d} pre-trained on K400 dataset \cite{k400}; (2) TimeSformer(TSF) \cite{timesformer} pre-trained on K600 \cite{k600} and Ego-Exo4D(EE4D) \cite{egoexo4d} datasets; (3) EgoVLP \cite{egovlp} pre-trained on Ego4D(E4D) \cite{ego4d} dataset.

\begin{table}[t]
    \caption{\textbf{Action classification benchmark results on different models}. Top-1 accuracies are reported for different models with different pretraining strategies. \textbf{Bolded} and \ul{underlined} values indicates the best and 2-nd best results, respectively.}
    \label{tab:action_recognition_methods}
    \vspace{-0.2cm}
    \centering
    \resizebox{0.9\linewidth}{!}{

    \begin{tabular}{c|c|c|cc|c|c|c|c|cc}
        \toprule  
        \multirow{2}{*}{\textbf{Train on}} & \multirow{2}{*}{\textbf{Models}} & \multirow{2}{*}{\textbf{Pretrain}} & \multicolumn{2}{c|}{\textbf{Test on}} & & \multirow{2}{*}{\textbf{Train on}} & \multirow{2}{*}{\textbf{Models}} & \multirow{2}{*}{\textbf{Pretrain}} & \multicolumn{2}{c}{\textbf{Test on}} \\
        \multirow{2}{*}{} & \multirow{2}{*}{}  & {} & {Exo} & {Ego} & & \multirow{2}{*}{} & \multirow{2}{*}{}  & \multirow{2}{*}{} & {Exo} & {Ego}\\
        \midrule
        Exo &       &  & \ul{0.9194} & 0.0927 & & Exo         &  & & 0.8940 & 0.0893 \\
        Ego &       I3D\cite{i3d} & K400\cite{k400} & 0.1025 & 0.7469  & & Ego         & EgoVLP\cite{egovlp} & Ego4D\cite{ego4d} & 0.0887 & \ul{0.7977}\\
        Ego \& Exo  & &  & 0.8963 & 0.7266 & & Ego \& Exo  &   & & 0.8986 & 0.7932 \\
        \midrule
        Exo         & &  & \textbf{0.9274} & 0.0836  & & Exo         & & K600\cite{k600} & 0.8825 & 0.0814   \\
        Ego         & TSF\cite{timesformer} & K600\cite{k600} & 0.1417 & 0.7932  & & Ego         & TSF\cite{timesformer} & + & 0.1601 & \textbf{0.8000}\\
        Ego \& Exo  &  &  & 0.8894 & 0.7842  & & Ego \& Exo  &  & Ego-Exo4D\cite{egoexo4d} & 0.8975 & 0.7840 \\
    \bottomrule  
    \end{tabular}
    \vspace{-0.3cm}
}
\end{table}

\vspace{0.1cm}
\noindent{\textbf{Experiment Results.}}
Top-1 accuracies of different models are reported in \cref{tab:action_recognition_methods}. We analyze the results in the following aspects.
\textbf{(1) \emph{Impacts of pretraining}}. Among all results, TSF and I3D pre-trained on Kinetics datasets achieve the best two performances (0.9274 and 0.9194) on exocentric videos. Similarly, TSF pre-trained on EE4D performs best (0.8000) on egocentric videos, closely followed by the one pre-trained on E4D (0.7977). Such results are attributed to view-related pre-training datasets  (\ie, Kinetics are exocentric datasets; E4D and EE4D consist of various egocentric videos).
\textbf{(2) \emph{Analysis on the view gap.}} 
Not surprisingly, models trained ego-only or exo-only data suffer from a significant performance drop on cross-view testing.
Additionally, we find that mixing up cross-view data (Ego \& Exo) for training does not always bring performance improvement. For I3D and TSF pre-trained on Kinetics datasets, performance drops on both egocentric and exocentric data. For TSF pre-trained from E4D and EE4D, only performance on exocentric data obtains improvement when mixing up cross-view data for training.
Such results indicate a great domain gap between ego-videos and exo-videos.
\textbf{(3) \emph{Why do models perform worse on ego-videos?}}
From \cref{tab:action_recognition_methods}, we also observe that models always perform worse on ego-videos than on exo-videos.
We think this is because it is easier to observe similar action patterns from egocentric videos, which confuse models. Another reason is that it is more difficult to find discriminating clues from the Ego-M camera. Appendix A3.1 provides more analysis supporting these views.

\vspace{-0.2cm}
\subsection{Cross-view Sequence Verification}
\label{sec:cvsv}
\vspace{-0.1cm}

Sequence Verification (SV) \cite{svip, weakly-svr,he2023collaborative} is proposed to verify the action order consistency of sequential videos under a scenario where precise temporal annotations are not provided, which shows great potential in video abstraction, industrial safety, and skill studying. 
Existing SV datasets \cite{svip} are collected either from exocentric or egocentric cameras, which constrains existing studies in an inner-view manner.
However, it is desirable to study whether a model can perform promising verification of two videos from egocentric and exocentric views. 
For instance, during our daily fitness exercises, an AI assistant in our eyewear can remind us whether we have missed any exercise program by verifying the sequence of exocentric expertise exemplar videos and the self-recorded egocentric videos.  
Hence, we extend the traditional SV to Cross-View SV (CVSV).

\vspace{0.1cm}
\noindent{\textbf{Task Setups.}}
CVSV aims to verify whether two fitness sequence videos have identical procedures. Two action sequence videos executing the same steps in the same order form a positive pair; otherwise, they are negative. 
The method should give a verification distance between each video pair based on the video representations, and give the prediction by thresholding the distance. 

CVSV is more challenging than traditional SV because videos can be shot from either egocentric or exocentric cameras. Hence, it is crucial for models to learn retrievable (or translatable) representations across views. More formal task setups will be introduced in Appendix A3.2.

\vspace{0.1cm}
\noindent{\textbf{Baseline model.}} We use the state-of-the-art SV model CAT \cite{svip} to conduct experiments. More details about the CAT will be introduced in Appendix A3.2.

\vspace{0.1cm}
\noindent{\textbf{Metrics.}}
\label{sec:sv-matrics}
\textbf{(1) \emph{AUC}}: Following existing works \cite{svip,weakly-svr,he2023collaborative}, we first adopt the Area Under ROC Curve to evaluate the performance. 
\textbf{(2) \emph{Rank 1} \& \emph{mAP}}: To further study the relations among learned representations, we borrow the idea of image retrieval \cite{caron2021emerging,jang2021self,kim2021cds} and use Rank-1 and mAP to evaluate CVSV models. 

\vspace{0.1cm}
\noindent{\textbf{Experiment Results.}}
Benchmark results are reported in \cref{tab:sv_arrangement_all} and \cref{tab:sv_ego_rate}. We analyze the results in the following aspects.
\textbf{(1) \emph{Influence of cross-view training data.}} 
As the first attempt, we wonder how cross-view training data will influence the performance.
Hence, we separate all training video pairs into three parts (\ie, \emph{Exo-Exo}, \emph{Ego-Ego} and \emph{Exo-Ego}) to study how cross-view training data would influence the performance. The results are shown in \cref{tab:sv_arrangement_all}.
First, we observe that combining all training pairs will benefit performance on exo-only and ego-exo pairs but bring a performance drop on ego-only pairs, which further indicates the domain gap between different views. 
Furthermore, compared with 0.8033 on ego-only data and 0.8221 on exo-only data, the best SV performance on ego-exo data is 0.7755, which indicates that cross-view sequence verification is a challenging task. Retrieval results also support this conclusion. Cross-view retrieval achieves much poorer performance (0.3 on Rank 1 and 0.25 on mAP) compared with inner-view retrieval.
\textbf{(2) \emph{How many egocentric data is needed for CVSV?}} In practical application, it is much easier to collect exocentric videos than egocentric videos. Hence, it is desirable to study if a CVSV model can achieve superior performance with limited egocentric training videos. 
To the end, we gradually prune egocentric videos from the training set (\ie, 0\%, 30\%, 70\% and 100\%) and evaluate the performance. 
As shown in \cref{tab:sv_ego_rate}, when gradually prone training data of egocentric videos, the performance drops on all metrics, posing a great challenge for future study on settings with unbalanced (\ie, limited egocentric videos and rich exocentric videos) cross-view data.
\begin{table}[t]
    \caption{\textbf{Cross-view sequence verification results on various training sources.} We find that mixing-up all training pairs will benifit exo-only and ego-exo SV, but bring declines on ego-only setting. ``View1-View2'' indicates that the model takes video pairs with one video from \emph{View1} and the other from \emph{View2}. ``View1$\rightarrow$View2'' indicates taking videos from \emph{View1} to retrieve videos from \emph{View2}. }
    \vspace{-0.2cm}
    \label{tab:sv_arrangement_all}
    \scriptsize
        \centering
        \scriptsize
        \resizebox{\linewidth}{!}{
        \begin{tabular}{ccc|ccc|cccc|cccc}
        \toprule
        \multicolumn{3}{c|}{\textbf{Train on}} & \multicolumn{3}{c|}{\textbf{AUC}} & \multicolumn{4}{c|}{\textbf{Rank 1}} & \multicolumn{4}{c}{\textbf{mAP}}\\  
        Exo-Exo & Ego-Ego & Exo-Ego & Ego-Ego & Exo-Exo & Ego-Exo & 
        Ego$\rightarrow$Ego &
        Exo$\rightarrow$Exo &
        Ego$\rightarrow$Exo &
        Exo$\rightarrow$Ego &
        Ego$\rightarrow$Ego &
        Exo$\rightarrow$Exo &
        Ego$\rightarrow$Exo &
        Exo$\rightarrow$Ego \\
        \midrule
        \checkmark&&                    & 0.532 & 0.800 & 0.577                  & 0.165 & 0.583 & 0.094 & 0.092                        & 0.087 & 0.374 & 0.117 & 0.080 \\
        &\checkmark&                    & \textbf{0.803} & 0.487 & 0.480         & \textbf{0.620} & 0.071 & 0.040 & 0.021               & \textbf{0.325} & 0.064 & 0.057 & 0.065 \\
        &&\checkmark                    & 0.761 & 0.813 & 0.744                  & 0.539 & 0.646 & 0.296 & 0.363                        & 0.275 & 0.394 & 0.228 & 0.237 \\
        \checkmark&\checkmark&          & 0.751 & 0.814 & 0.743                  & 0.556 & 0.629 & 0.286 & \textbf{0.383}               & 0.275 & 0.402 & 0.223 & 0.238 \\
        \checkmark&\checkmark&\checkmark& 0.759 & \textbf{0.822} & \textbf{0.776}& 0.572 & \textbf{0.663} & \textbf{0.300} & 0.367      & 0.281 & \textbf{0.406} & \textbf{0.247} & \textbf{0.247} \\
        \bottomrule
        \end{tabular}
        }
\end{table}

\begin{table}[t]
    \caption{\textbf{Cross-view sequence verification results on imbalanced training data}. We gradually prune egocentric videos from training data and we find that it is challenging to perform cross-view sequence verification with imbalanced data.}
    \vspace{-0.2cm}
    \label{tab:sv_ego_rate}
    \scriptsize
        \centering
        \scriptsize
        \resizebox{1\linewidth}{!}{
        \begin{tabular}{c|c|cc|cc||c|c|cc|cc}
            \toprule
            {\textbf{Prune}}& \textbf{AUC} &\multicolumn{2}{c|}{\textbf{Rank 1}} & \multicolumn{2}{c||}{\textbf{mAP}} & {\textbf{Prune}}& \textbf{AUC} &\multicolumn{2}{c|}{\textbf{Rank 1}} & \multicolumn{2}{c}{\textbf{mAP}}\\ 
            \textbf{Rate} & Exo-Ego &
            Ego$\rightarrow$Exo &
            Exo$\rightarrow$Ego &
            Ego$\rightarrow$Exo &
            Exo$\rightarrow$Ego &
            \textbf{Rate} & Exo-Ego &
            Ego$\rightarrow$Exo &
            Exo$\rightarrow$Ego &
            Ego$\rightarrow$Exo &
            Exo$\rightarrow$Ego \\
            \midrule
            100\% & 0.5768 & 0.0943 & 0.0917 & 0.1174 & 0.0798 & 30\% & 0.7072 & 0.1751 & 0.2917 & 0.1824 & 0.1905 \\
            70\% & 0.6562 & 0.1077 & 0.0750 & 0.1412 & 0.1109 & 0\% & \textbf{0.7755} & \textbf{0.2997} & \textbf{0.3667} & \textbf{0.2470} & \textbf{0.2468} \\

            \bottomrule
        \end{tabular}
        }
        \vspace{-0.3cm}
\end{table}

\vspace{-0.2cm}
\subsection{Guidance-based Execution Verification}
\label{sec:gev}
\vspace{-0.1cm}
Existing works in Action Assessment mainly focus on predicting the final score of an action video or a pair-wise ranking between a pair of videos. 
However, in real-world action guiding scenarios, providing interpretable feedback is more valuable than giving a score or a ranking. 
For example, our fitness coach will tell us which technical keypoints are not satisfied in our executions, which will not only explain how well we have performed but also let us know how to improve. 
However, such an ability has never been explored in action assessment.
To address this issue, we make the first attempt to study interpretable action assessment and propose a novel task termed Guidance-based Execution Verification (GEV).

\vspace{0.1cm}
\noindent{\textbf{Task Setups}.}
Given a set of technical keypoints in text as the guidance, the goal of GEV is to verify whether the execution of an action satisfies the keypoints in the guidance.
Formally, given an action video $v$ and $n$ technical keypoints $Q=\{q_1, q_2,..., q_n\}$, a model $F$ is asked to perform an $n$-way score prediction $P=\{p_1,...,p_n\}=F(v,Q)$, where $p_i$ represents the verification score of the $i$-th keypoint. The higher the $p_i$, the more likely the action satisfies the $i$-th keypoint.
During inference, a threshold $\tau$ is adopted to verify whether the action satisfies the keypoints. If $p_i$ > $\tau$, the model predicts the action ``satisfies'' the $i$-th keypoint. Otherwise, the model returns a result of being ``unsatisfies''.

\begin{figure}[t]
  \centering
  \includegraphics[width=1\linewidth]{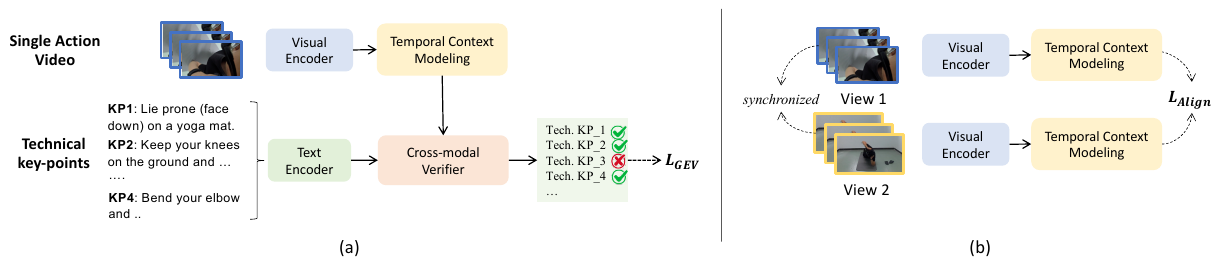}
  \caption{\textbf{Overview of GEVFormer}. (a) GEVFormer takes an action video and technical keypoints as input, and output the verification results on each keypoint. (b) During training, a synchronized video alignment loss is adopted to force the model to obtain consistent representations across synchronized videos from various views.}
  \label{fig:gevformer}
  \vspace{-0.3cm}
\end{figure}

\vspace{0.1cm}
\noindent{\textbf{Baseline Model.}} To better address GEV, 
we introduce a transformer-based \cite{transformer} model named GEVFormer, which tasks a single action video and the corresponding technical keypoints as input and outputs the verification results for each keypoint. As shown in \cref{fig:gevformer}(a), video and keypoints are fed into the visual and text encoder to obtain visual and text embeddings. After that, a Temporal Context Modeling (TCM) module is adopted to model temporal information of visual embeddings further, obtaining enhanced visual embeddings. 
Finally, text embeddings and enhanced visual embeddings are fed into a Cross-Modal Verifier (CMV) to obtain the results.

During training, a loss for GEV, denoted as $L_{GEV}$, is adopted to require the model to provide accurate verification results. Besides, in our early experimental attempts on GEV, we have the same observations as in other tasks that simply combining training data from egocentric and exocentric views cannot bring stable performance improvement due to the domain gap between different views. To bridge the gap, as shown in \cref{fig:gevformer}(b), inspired by previous works on cross-view learning \cite{yu2019see,tian2020contrastive}, we propose to utilize an InfoNCE-based \cite{infonce} alignment loss, denoted as $L_{Align}$, to force model to obtain consistent representations across synchronized videos from various views. The overall training loss is written as $L=L_{GEV}+\lambda L_{Align}$, where $\lambda$ is a hyper-parameter.

In our implementation, visual and text encoders are designed as the image and text encoders of pre-trained CLIP \cite{clip} and frozen during training. The TCM is designed as a Transformer Encoder and the CMV contains a Transformer Decoder together with a linear evaluator. See more details in Appendix A3.3.

\begin{table}[t]
    \caption{\textbf{Guidance-based execution verification results on different baselines}. We report performance on ego- and exo-centric data alone with the average F1-score. ``ego/exo'' indicates ndependent models are trained on ego-only and exo-only data. ``ego+exo'' indicates the model is trained on both ego-and-exo data.}
    \vspace{-0.2cm}
    \label{tab:execution_verification_clip}
    \scriptsize
        \centering
        \scriptsize
        \resizebox{0.9\linewidth}{!}{
        \begin{tabular}{l|ccc|ccc|c}
        \toprule
        \multirow{2}{*}{\textbf{Methods}} & \multicolumn{3}{c|}{\textbf{Exo}}  & \multicolumn{3}{c|}{\textbf{Ego}} & \textbf{Avg}\\
        \multirow{2}{*}{} & F1-score & Precision & Recall & F1-score & Precision & Recall & F1-score\\
        \midrule
        Random                          & 0.3178          & \textcolor{lightgray}{0.2329} & \textcolor{lightgray}{0.5000} & 0.3178          & \textcolor{lightgray}{0.2329} & \textcolor{lightgray}{0.5000}                        & 0.3178 \\
        Distribution Prior              & 0.2323          & \textcolor{lightgray}{0.2323} & \textcolor{lightgray}{0.2323} & 0.2323          & \textcolor{lightgray}{0.2323} & \textcolor{lightgray}{0.2323}                        & 0.2323\\
        CLIP-GEV(ego+exo)            & 0.5080          & \textcolor{lightgray}{0.5362} & \textcolor{lightgray}{0.4657} & 0.4780          & \textcolor{lightgray}{0.5401} & \textcolor{lightgray}{0.4094} & 0.4881      \\
        \midrule
        GEVFormer(ego/exo) w/o alignment & \textbf{0.5474} & \textcolor{lightgray}{0.5541} & \textcolor{lightgray}{0.5408} & 0.5161          & \textcolor{lightgray}{0.5067} & \textcolor{lightgray}{0.5259} & 0.5318\\ 
        GEVFormer(ego+exo) w/o alignment & 0.5282          & \textcolor{lightgray}{0.5502} & \textcolor{lightgray}{0.5080} & \ul{0.5248}     & \textcolor{lightgray}{0.5570} & \textcolor{lightgray}{0.4960} & \ul{0.5265} \\
        \midrule  
        \rowcolor{yellow} GEVFormer(ego+exo)  & \ul{0.5452}     & \textcolor{lightgray}{0.5219} & \textcolor{lightgray}{0.5707} & \textbf{0.5425} & \textcolor{lightgray}{0.5186} & \textcolor{lightgray}{0.5687} & \textbf{0.5439}\\     
        \bottomrule
        \end{tabular}
        }
        \vspace{-0.3cm}
\end{table}

\vspace{0.1cm}
\noindent{\textbf{Experiment Results.}} 
We compare GEVFormer with four other naive methods and invariants: \textbf{(1) \emph{Random}}: Randomly predict an action satisfies a technical keypoint with 50\% probability; \textbf{(2) \emph{Distribution Prior}}: Randomly predict an action satisfies a keypoint with the distribution prior;  
\textbf{(3) \emph{CLIP-GEV}}: Simply concatenate average-pooled visual embedding and text embeddings extracted by CLIP \cite{clip} and feed it into a linear evaluator to predict the results; \textbf{(4) \emph{GEVFormer w/o alignment}}: Ablate $L_{Align}$ from GEVFormer. 
To evaluate the methods, we adopt the \emph{Precision}, \emph{Recall}, and \emph{F1-score}. Here we regard ``unsatisfies'' as the positive label since samples with ``satisfies'' labels take up a much higher proportion than those with ``unsatisfies'' labels in EgoExo-Fitness.

As shown in \cref{tab:execution_verification_clip}, GEVFormer outperforms all naive baselines. Besides, compared with the variants of GEVFormer, we have the same findings as on the other tasks that jointly training models on ego- and exo-centric data will not bring stable improvement (achieving 0.0087 improvement on egocxentric data with 0.0192 drop on exocentric data). 
Surprisingly, when further adopting the $L_{Align}$, GEVFormer achieves the best performance on egocentric data with 0.5425 F1-score, suffering only a 0.0022 performance drop on exocentric data.

\vspace{-0.2cm}
\section{Conclusion}
\vspace{-0.1cm}
We believe that studying egocentric full-body action understanding will benefit the development of AI-assistant. To enable this line of research, we focus on the scenario of fitness exercise and guiding, and introduce EgoExo-Fitness.
With a diverse range of synchronized ego- and exo-centric fitness action sequence videos and rich annotations on temporal boundaries and interpretable action judgement, EgoExo-Fitness provides new resources for egocentric and exocentric full-body action understanding.
To facilitate future research on EgoExo-Fitness, we construct benchmarks on five relevant tasks. 
Through experiment analysis, we evaluate the performance of baseline models and point out several interesting problems that await future research (\eg, how to better address cross-view modeling with unbalanced data; how to leverage synchronized exocentric data to achieve better performance).

\section*{Acknowledgments}
This work was supported partially by the National Key Research and Development Program of China (2023YFA1008503), NSFC(U21A20471, 62206315), Guangdong NSF Project (No. 2023B1515040025, 2020B1515120085, 2024A15150-10101), Guangzhou Basic and Applied Basic Research Scheme(2024A04J4067). 
The authors thank Kun-Yu Lin for the valuable discussions. The authors also thank anonymous reviewers and ACs for their constructive suggestions.

%
%
\bibliographystyle{splncs04}
\bibliography{main}

\begin{thebibliography}{10}
\providecommand{\url}[1]{\texttt{#1}}
\providecommand{\urlprefix}{URL }
\providecommand{\doi}[1]{https://doi.org/#1}

\bibitem{unrealego}
Akada, H., Wang, J., Shimada, S., Takahashi, M., Theobalt, C., Golyanik, V.: Unrealego: A new dataset for robust egocentric 3d human motion capture. In: Proceedings of the European Conference on Computer Vision. pp. 1--17. Springer (2022)

\bibitem{ego_procel}
Bansal, S., Arora, C., Jawahar, C.: My view is the best view: Procedure learning from egocentric videos. In: Proceedings of the European Conference on Computer Vision. pp. 657--675. Springer (2022)

\bibitem{1st-basketball}
Bertasius, G., Soo~Park, H., Yu, S.X., Shi, J.: Am i a baller? basketball performance assessment from first-person videos. In: Proceedings of the IEEE/CVF International Conference on Computer Vision. pp. 2177--2185 (2017)

\bibitem{timesformer}
Bertasius, G., Wang, H., Torresani, L.: Is space-time attention all you need for video understanding? In: Proceedings of the International Conference on Machine Learning (2021)

\bibitem{wear}
Bock, M., Moeller, M., Van~Laerhoven, K., Kuehne, H.: Wear: A multimodal dataset for wearable and egocentric video activity recognition. arXiv preprint arXiv:2304.05088  (2023)

\bibitem{humman_eccv22}
Cai, Z., Ren, D., Zeng, A., Lin, Z., Yu, T., Wang, W., Fan, X., Gao, Y., Yu, Y., Pan, L., et~al.: Humman: Multi-modal 4d human dataset for versatile sensing and modeling. In: Proceedings of the European Conference on Computer Vision. pp. 557--577. Springer (2022)

\bibitem{caron2021emerging}
Caron, M., Touvron, H., Misra, I., J{\'e}gou, H., Mairal, J., Bojanowski, P., Joulin, A.: Emerging properties in self-supervised vision transformers. In: Proceedings of the IEEE/CVF International Conference on Computer Vision. pp. 9650--9660 (2021)

\bibitem{k600}
Carreira, J., Noland, E., Banki-Horvath, A., Hillier, C., Zisserman, A.: A short note about kinetics-600. arXiv preprint arXiv:1808.01340  (2018)

\bibitem{i3d}
Carreira, J., Zisserman, A.: Quo vadis, action recognition? a new model and the kinetics dataset. In: Proceedings of the IEEE/CVF Conference on Computer Vision and Pattern Recognition. pp. 6299--6308 (2017)

\bibitem{epic-kitchen}
Damen, D., Doughty, H., Farinella, G.M., Fidler, S., Furnari, A., Kazakos, E., Moltisanti, D., Munro, J., Perrett, T., Price, W., et~al.: Scaling egocentric vision: The epic-kitchens dataset. In: Proceedings of the European Conference on Computer Vision. pp. 720--736 (2018)

\bibitem{epic-kitchen-100}
Damen, D., Doughty, H., Farinella, G.M., Furnari, A., Kazakos, E., Ma, J., Moltisanti, D., Munro, J., Perrett, T., Price, W., et~al.: Rescaling egocentric vision. arXiv preprint arXiv:2006.13256  (2020)

\bibitem{actionsense}
DelPreto, J., Liu, C., Luo, Y., Foshey, M., Li, Y., Torralba, A., Matusik, W., Rus, D.: Actionsense: A multimodal dataset and recording framework for human activities using wearable sensors in a kitchen environment. Advances in Neural Information Processing Systems  \textbf{35},  13800--13813 (2022)

\bibitem{adl}
Diete, A., Sztyler, T., Stuckenschmidt, H.: Vision and acceleration modalities: Partners for recognizing complex activities. In: Proceedings of the IEEE International Conference on Pervasive Computing and Communications Workshops. pp. 101--106. IEEE (2019)

\bibitem{weakly-svr}
Dong, S., Hu, H., Lian, D., Luo, W., Qian, Y., Gao, S.: Weakly supervised video representation learning with unaligned text for sequential videos. In: Proceedings of the IEEE/CVF Conference on Computer Vision and Pattern Recognition. pp. 2437--2447 (2023)

\bibitem{easycom}
Donley, J., Tourbabin, V., Lee, J.S., Broyles, M., Jiang, H., Shen, J., Pantic, M., Ithapu, V.K., Mehra, R.: Easycom: An augmented reality dataset to support algorithms for easy communication in noisy environments. arXiv preprint arXiv:2107.04174  (2021)

\bibitem{whos_better}
Doughty, H., Damen, D., Mayol-Cuevas, W.: Who's better? who's best? pairwise deep ranking for skill determination. In: Proceedings of the IEEE/CVF Conference on Computer Vision and Pattern Recognition. pp. 6057--6066 (2018)

\bibitem{raan}
Doughty, H., Mayol-Cuevas, W., Damen, D.: The pros and cons: Rank-aware temporal attention for skill determination in long videos. In: Proceedings of the IEEE/CVF Conference on Computer Vision and Pattern Recognition. pp. 7862--7871 (2019)

\bibitem{du2022weakly}
Du, J.R., Feng, J.C., Lin, K.Y., Hong, F.T., Wu, X.M., Qi, Z., Shan, Y., Zheng, W.S.: Weakly-supervised temporal action localization by progressive complementary learning. arXiv preprint arXiv:2206.11011  (2022)

\bibitem{aifit}
Fieraru, M., Zanfir, M., Pirlea, S.C., Olaru, V., Sminchisescu, C.: Aifit: Automatic 3d human-interpretable feedback models for fitness training. In: Proceedings of the IEEE/CVF Conference on Computer Vision and Pattern Recognition. pp. 9919--9928 (2021)

\bibitem{gao2020asymmetric}
Gao, J., Zheng, W.S., Pan, J.H., Gao, C., Wang, Y., Zeng, W., Lai, J.: An asymmetric modeling for action assessment. In: Proceedings of the European Conference on Computer Vision. pp. 222--238. Springer (2020)

\bibitem{jigsaws}
Gao, Y., Vedula, S.S., Reiley, C.E., Ahmidi, N., Varadarajan, B., Lin, H.C., Tao, L., Zappella, L., B{\'e}jar, B., Yuh, D.D., et~al.: Jhu-isi gesture and skill assessment working set (jigsaws): A surgical activity dataset for human motion modeling. In: Proceedings of the International Conference on Medical Image Computing and Computer-Assisted Intervention Workshop (2014)

\bibitem{ata}
Ghoddoosian, R., Dwivedi, I., Agarwal, N., Dariush, B.: Weakly-supervised action segmentation and unseen error detection in anomalous instructional videos. In: Proceedings of the IEEE/CVF International Conference on Computer Vision. pp. 10128--10138 (2023)

\bibitem{ego4d}
Grauman, K., Westbury, A., Byrne, E., Chavis, Z., Furnari, A., Girdhar, R., Hamburger, J., Jiang, H., Liu, M., Liu, X., et~al.: Ego4d: Around the world in 3,000 hours of egocentric video. In: Proceedings of the IEEE/CVF Conference on Computer Vision and Pattern Recognition. pp. 18995--19012 (2022)

\bibitem{egoexo4d}
Grauman, K., Westbury, A., Torresani, L., Kitani, K., Malik, J., Afouras, T., Ashutosh, K., Baiyya, V., Bansal, S., Boote, B., et~al.: Ego-exo4d: Understanding skilled human activity from first-and third-person perspectives. In: Proceedings of the IEEE/CVF Conference on Computer Vision and Pattern Recognition. pp. 19383--19400 (2024)

\bibitem{hps}
Guzov, V., Mir, A., Sattler, T., Pons-Moll, G.: Human poseitioning system (hps): 3d human pose estimation and self-localization in large scenes from body-mounted sensors. In: Proceedings of the IEEE/CVF Conference on Computer Vision and Pattern Recognition. pp. 4318--4329 (2021)

\bibitem{he2023collaborative}
He, T., Liu, H., Li, Y., Ma, X., Zhong, C., Zhang, Y., Lin, W.: Collaborative weakly supervised video correlation learning for procedure-aware instructional video analysis. arXiv preprint arXiv:2312.11024  (2023)

\bibitem{repcount}
Hu, H., Dong, S., Zhao, Y., Lian, D., Li, Z., Gao, S.: Transrac: Encoding multi-scale temporal correlation with transformers for repetitive action counting. In: Proceedings of the IEEE/CVF Conference on Computer Vision and Pattern Recognition. pp. 19013--19022 (2022)

\bibitem{egoexolearn}
Huang, Y., Chen, G., Xu, J., Zhang, M., Yang, L., Pei, B., Zhang, H., Dong, L., Wang, Y., Wang, L., et~al.: Egoexolearn: A dataset for bridging asynchronous ego-and exo-centric view of procedural activities in real world. In: Proceedings of the IEEE/CVF Conference on Computer Vision and Pattern Recognition. pp. 22072--22086 (2024)

\bibitem{thumos}
Idrees, H., Zamir, A.R., Jiang, Y.G., Gorban, A., Laptev, I., Sukthankar, R., Shah, M.: The thumos challenge on action recognition for videos “in the wild”. Computer Vision and Image Understanding  \textbf{155},  1--23 (2017)

\bibitem{human36m}
Ionescu, C., Papava, D., Olaru, V., Sminchisescu, C.: Human3. 6m: Large scale datasets and predictive methods for 3d human sensing in natural environments. IEEE Transactions on Pattern Analysis and Machine Intelligence  \textbf{36}(7),  1325--1339 (2013)

\bibitem{jang2021self}
Jang, Y.K., Cho, N.I.: Self-supervised product quantization for deep unsupervised image retrieval. In: Proceedings of the IEEE/CVF International Conference on Computer Vision. pp. 12085--12094 (2021)

\bibitem{epic-tent}
Jang, Y., Sullivan, B., Ludwig, C., Gilchrist, I., Damen, D., Mayol-Cuevas, W.: Epic-tent: An egocentric video dataset for camping tent assembly. In: Proceedings of the IEEE/CVF International Conference on Computer Vision Workshops. pp.~0--0 (2019)

\bibitem{k400}
Kay, W., Carreira, J., Simonyan, K., Zhang, B., Hillier, C., Vijayanarasimhan, S., Viola, F., Green, T., Back, T., Natsev, P., et~al.: The kinetics human action video dataset. arXiv preprint arXiv:1705.06950  (2017)

\bibitem{egohumans}
Khirodkar, R., Bansal, A., Ma, L., Newcombe, R., Vo, M., Kitani, K.: Egohumans: An egocentric 3d multi-human benchmark. arXiv preprint arXiv:2305.16487  (2023)

\bibitem{kim2021cds}
Kim, D., Saito, K., Oh, T.H., Plummer, B.A., Sclaroff, S., Saenko, K.: Cds: Cross-domain self-supervised pre-training. In: Proceedings of the IEEE/CVF International Conference on Computer Vision. pp. 9123--9132 (2021)

\bibitem{action_recog_survey}
Kong, Y., Fu, Y.: Human action recognition and prediction: A survey. International Journal of Computer Vision  (2022)

\bibitem{h2o}
Kwon, T., Tekin, B., St{\"u}hmer, J., Bogo, F., Pollefeys, M.: H2o: Two hands manipulating objects for first person interaction recognition. In: Proceedings of the IEEE/CVF International Conference on Computer Vision. pp. 10138--10148 (2021)

\bibitem{assistsr}
Lei, S.W., Gao, D., Wang, Y., Mao, D., Liang, Z., Ran, L., Shou, M.Z.: Assistsr: Task-oriented video segment retrieval for personal ai assistant. arXiv preprint arXiv:2111.15050  (2021)

\bibitem{egoego}
Li, J., Liu, K., Wu, J.: Ego-body pose estimation via ego-head pose estimation. In: Proceedings of the IEEE/CVF Conference on Computer Vision and Pattern Recognition. pp. 17142--17151 (2023)

\bibitem{egtea}
Li, Y., Liu, M., Rehg, J.M.: In the eye of beholder: Joint learning of gaze and actions in first person video. In: Proceedings of the European conference on computer vision (ECCV). pp. 619--635 (2018)

\bibitem{diving48}
Li, Y., Li, Y., Vasconcelos, N.: Resound: Towards action recognition without representation bias. In: Proceedings of the European Conference on Computer Vision (ECCV). pp. 513--528 (2018)

\bibitem{con_aqa}
Li, Y.M., Zeng, L.A., Meng, J.K., Zheng, W.S.: Continual action assessment via task-consistent score-discriminative feature distribution modeling. IEEE Transactions on Circuits and Systems for Video Technology  (2024)

\bibitem{egovlp}
Lin, K.Q., Wang, J., Soldan, M., Wray, M., Yan, R., XU, E.Z., Gao, D., Tu, R.C., Zhao, W., Kong, W., et~al.: Egocentric video-language pretraining. Advances in Neural Information Processing Systems  \textbf{35},  7575--7586 (2022)

\bibitem{lin2024rethinking}
Lin, K.Y., Ding, H., Zhou, J., Peng, Y.X., Zhao, Z., Loy, C.C., Zheng, W.S.: Rethinking clip-based video learners in cross-domain open-vocabulary action recognition. arXiv preprint arXiv:2403.01560  (2024)

\bibitem{bmn}
Lin, T., Liu, X., Li, X., Ding, E., Wen, S.: Bmn: Boundary-matching network for temporal action proposal generation. In: Proceedings of the IEEE/CVF International Conference on Computer Vision. pp. 3889--3898 (2019)

\bibitem{ntu-120}
Liu, J., Shahroudy, A., Perez, M., Wang, G., Duan, L.Y., Kot, A.C.: Ntu rgb+ d 120: A large-scale benchmark for 3d human activity understanding. IEEE Transactions on Pattern Analysis and Machine Intelligence  \textbf{42}(10),  2684--2701 (2019)

\bibitem{tadtr}
Liu, X., Wang, Q., Hu, Y., Tang, X., Zhang, S., Bai, S., Bai, X.: End-to-end temporal action detection with transformer. IEEE Transactions on Image Processing  \textbf{31},  5427--5441 (2022)

\bibitem{hoi4d}
Liu, Y., Liu, Y., Jiang, C., Lyu, K., Wan, W., Shen, H., Liang, B., Fu, Z., Wang, H., Yi, L.: Hoi4d: A 4d egocentric dataset for category-level human-object interaction. In: Proceedings of the IEEE/CVF Conference on Computer Vision and Pattern Recognition. pp. 21013--21022 (2022)

\bibitem{you2me}
Ng, E., Xiang, D., Joo, H., Grauman, K.: You2me: Inferring body pose in egocentric video via first and second person interactions. In: Proceedings of the IEEE/CVF Conference on Computer Vision and Pattern Recognition. pp. 9890--9900 (2020)

\bibitem{infonce}
Oord, A.v.d., Li, Y., Vinyals, O.: Representation learning with contrastive predictive coding. arXiv preprint arXiv:1807.03748  (2018)

\bibitem{aqa7}
Parmar, P., Morris, B.: Action quality assessment across multiple actions. In: Proceedings of the IEEE Winter Conference on Applications of Computer Vision. pp. 1468--1476. IEEE (2019)

\bibitem{mtl-aqa}
Parmar, P., Morris, B.T.: What and how well you performed? a multitask learning approach to action quality assessment. In: Proceedings of the IEEE/CVF Conference on Computer Vision and Pattern Recognition. pp. 304--313 (2019)

\bibitem{outlook_ego}
Plizzari, C., Goletto, G., Furnari, A., Bansal, S., Ragusa, F., Farinella, G.M., Damen, D., Tommasi, T.: An outlook into the future of egocentric vision. arXiv preprint arXiv:2308.07123  (2023)

\bibitem{dataego}
Possas, R., Caceres, S.P., Ramos, F.: Egocentric activity recognition on a budget. In: Proceedings of the IEEE Conference on Computer Vision and Pattern Recognition. pp. 5967--5976 (2018)

\bibitem{svip}
Qian, Y., Luo, W., Lian, D., Tang, X., Zhao, P., Gao, S.: Svip: Sequence verification for procedures in videos. In: Proceedings of the IEEE/CVF Conference on Computer Vision and Pattern Recognition. pp. 19890--19902 (2022)

\bibitem{clip}
Radford, A., Kim, J.W., Hallacy, C., Ramesh, A., Goh, G., Agarwal, S., Sastry, G., Askell, A., Mishkin, P., Clark, J., et~al.: Learning transferable visual models from natural language supervision. In: Proceedings of the International Conference on Machine Learning. pp. 8748--8763. PMLR (2021)

\bibitem{meccano}
Ragusa, F., Furnari, A., Livatino, S., Farinella, G.M.: The meccano dataset: Understanding human-object interactions from egocentric videos in an industrial-like domain. In: Proceedings of the IEEE/CVF Winter Conference on Applications of Computer Vision. pp. 1569--1578 (2021)

\bibitem{ryan2023egocentric}
Ryan, F., Jiang, H., Shukla, A., Rehg, J.M., Ithapu, V.K.: Egocentric auditory attention localization in conversations. In: Proceedings of the IEEE/CVF Conference on Computer Vision and Pattern Recognition. pp. 14663--14674 (2023)

\bibitem{assembly101}
Sener, F., Chatterjee, D., Shelepov, D., He, K., Singhania, D., Wang, R., Yao, A.: Assembly101: A large-scale multi-view video dataset for understanding procedural activities. In: Proceedings of the IEEE/CVF Conference on Computer Vision and Pattern Recognition. pp. 21096--21106 (2022)

\bibitem{ntu-60}
Shahroudy, A., Liu, J., Ng, T.T., Wang, G.: Ntu rgb+ d: A large scale dataset for 3d human activity analysis. In: Proceedings of the IEEE/CVF Conference on Computer Vision and Pattern Recognition. pp. 1010--1019 (2016)

\bibitem{finegym}
Shao, D., Zhao, Y., Dai, B., Lin, D.: Finegym: A hierarchical video dataset for fine-grained action understanding. In: Proceedings of the IEEE/CVF Conference on Computer Vision and Pattern Recognition (2020)

\bibitem{charadesEgo}
Sigurdsson, G.A., Gupta, A., Schmid, C., Farhadi, A., Alahari, K.: Charades-ego: A large-scale dataset of paired third and first person videos. arXiv preprint arXiv:1804.09626  (2018)

\bibitem{coin}
Tang, Y., Ding, D., Rao, Y., Zheng, Y., Zhang, D., Zhao, L., Lu, J., Zhou, J.: Coin: A large-scale dataset for comprehensive instructional video analysis. In: Proceedings of the IEEE/CVF Conference on Computer Vision and Pattern Recognition. pp. 1207--1216 (2019)

\bibitem{flag3d}
Tang, Y., Liu, J., Liu, A., Yang, B., Dai, W., Rao, Y., Lu, J., Zhou, J., Li, X.: Flag3d: A 3d fitness activity dataset with language instruction. In: CVPR (2023)

\bibitem{tian2020contrastive}
Tian, Y., Krishnan, D., Isola, P.: Contrastive multiview coding. In: Proceedings of the European Conference on Computer Vision. pp. 776--794. Springer (2020)

\bibitem{cmu-mmac}
De~la Torre, F., Hodgins, J., Bargteil, A., Martin, X., Macey, J., Collado, A., Beltran, P.: Guide to the carnegie mellon university multimodal activity (cmu-mmac) database. Tech. report CMU-RI-TR-08-22, Robotics Institute, Carnegie Mellon University  (2009)

\bibitem{transformer}
Vaswani, A., Shazeer, N., Parmar, N., Uszkoreit, J., Jones, L., Gomez, A.N., Kaiser, {\L}., Polosukhin, I.: Attention is all you need. Advances in Neural Information Processing Systems  \textbf{30} (2017)

\bibitem{tal_survey}
Wang, B., Zhao, Y., Yang, L., Long, T., Li, X.: Temporal action localization in the deep learning era: A survey. IEEE Transactions on Pattern Analysis and Machine Intelligence  (2023)

\bibitem{egoscene}
Wang, J., Luvizon, D., Xu, W., Liu, L., Sarkar, K., Theobalt, C.: Scene-aware egocentric 3d human pose estimation. In: Proceedings of the IEEE/CVF Conference on Computer Vision and Pattern Recognition. pp. 13031--13040 (2023)

\bibitem{holoassist}
Wang, X., Kwon, T., Rad, M., Pan, B., Chakraborty, I., Andrist, S., Bohus, D., Feniello, A., Tekin, B., Frujeri, F.V., et~al.: Holoassist: an egocentric human interaction dataset for interactive ai assistants in the real world. In: Proceedings of the IEEE/CVF International Conference on Computer Vision. pp. 20270--20281 (2023)

\bibitem{assistq}
Wong, B., Chen, J., Wu, Y., Lei, S.W., Mao, D., Gao, D., Shou, M.Z.: Assistq: Affordance-centric question-driven task completion for egocentric assistant. In: Proceedings of the European Conference on Computer Vision. pp. 485--501. Springer (2022)

\bibitem{skating}
Xu, C., Fu, Y., Zhang, B., Chen, Z., Jiang, Y.G., Xue, X.: Learning to score figure skating sport videos. IEEE Transactions on Circuits and Systems for Video Technology  \textbf{30}(12),  4578--4590 (2019)

\bibitem{finediving}
Xu, J., Rao, Y., Yu, X., Chen, G., Zhou, J., Lu, J.: Finediving: A fine-grained dataset for procedure-aware action quality assessment. In: Proceedings of the IEEE/CVF Conference on Computer Vision and Pattern Recognition. pp. 2949--2958 (2022)

\bibitem{mo2cap2}
Xu, W., Chatterjee, A., Zollhoefer, M., Rhodin, H., Fua, P., Seidel, H.P., Theobalt, C.: Mo 2 cap 2: Real-time mobile 3d motion capture with a cap-mounted fisheye camera. IEEE Transactions on Visualization and Computer Graphics  \textbf{25}(5),  2093--2101 (2019)

\bibitem{yu2019see}
Yu, H., Cai, M., Liu, Y., Lu, F.: What i see is what you see: Joint attention learning for first and third person video co-analysis. In: Proceedings of the 27th ACM International Conference on Multimedia. pp. 1358--1366 (2019)

\bibitem{zeng2020hybrid}
Zeng, L.A., Hong, F.T., Zheng, W.S., Yu, Q.Z., Zeng, W., Wang, Y.W., Lai, J.H.: Hybrid dynamic-static context-aware attention network for action assessment in long videos. In: Proceedings of the ACM International Conference on Multimedia. pp. 2526--2534 (2020)

\bibitem{zeng2024multimodal}
Zeng, L.A., Zheng, W.S.: Multimodal action quality assessment. IEEE Transactions on Image Processing  (2024)

\bibitem{actionformer}
Zhang, C.L., Wu, J., Li, Y.: Actionformer: Localizing moments of actions with transformers. In: European Conference on Computer Vision (2022)

\bibitem{logo}
Zhang, S., Dai, W., Wang, S., Shen, X., Lu, J., Zhou, J., Tang, Y.: Logo: A long-form video dataset for group action quality assessment. In: Proceedings of the IEEE/CVF Conference on Computer Vision and Pattern Recognition. pp. 2405--2414 (2023)

\bibitem{egobody}
Zhang, S., Ma, Q., Zhang, Y., Qian, Z., Kwon, T., Pollefeys, M., Bogo, F., Tang, S.: Egobody: Human body shape and motion of interacting people from head-mounted devices. In: Proceedings of the European Conference on Computer Vision. pp. 180--200. Springer (2022)

\end{thebibliography}

\clearpage

\appendix

\setcounter{section}{0}
\setcounter{equation}{0}
\setcounter{figure}{0}
\setcounter{table}{0}

\renewcommand{\thesection}{A\arabic{section}}
\renewcommand{\thefigure}{A.\arabic{figure}}
\renewcommand{\thetable}{A.\arabic{table}}
\renewcommand{\theequation}{A.\arabic{equation}}

\section{Appendix}
In this Appendix, we will provide more details about data collection and annotations of the proposed EgoExo-Fitness dataset in \cref{sec:more_details_egoexo_fitness}. After that, we will introduce details about the benchmarks in \cref{sec:more_benchmark}, including formal definition, implementation, more experiment analysis, and other benchmark tasks. Finally, we provide more dicussinos about comparisons between EgoExo-Fitness and the existing datasets (\ie, Ego4D \cite{ego4d}, Ego-Exo4D \cite{egoexo4d}, and other related datasets) in \cref{sec:more_comparison_datasets}.

\section{More details of EgoExo-Fitness}
\label{sec:more_details_egoexo_fitness}
\noindent\textbf{Recording System.}
For the two Insta-Go3 cameras, we use $2560\times1440$ pixel resolution RGB images. For the GoPro camera, we use $1920\times1080$ pixel resolution RGB images. For the side and front exocentric cameras, we set the resolution to be $1024\times576$ and $1280\times720$, respectively. 
After synchronization, video frames will be extracted with 30 FPS and resized to $456\times256$.

\vspace{0.1cm}
\noindent\textbf{Participants.} We recruited 40 adults (28 males, 12 females) for data collection. Each participant was asked to participate at most nine rounds of recordings.

\vspace{0.1cm}
\noindent\textbf{Action sequences.}
EgoExo-Fitness records 86 types of fitness action sequences, each containing 3 to 6 continuous fitness actions. \cref{tab:sequence_list} provides the details of each action sequence.

\begin{figure}[ht]
  \centering
  \includegraphics[width=1\linewidth]{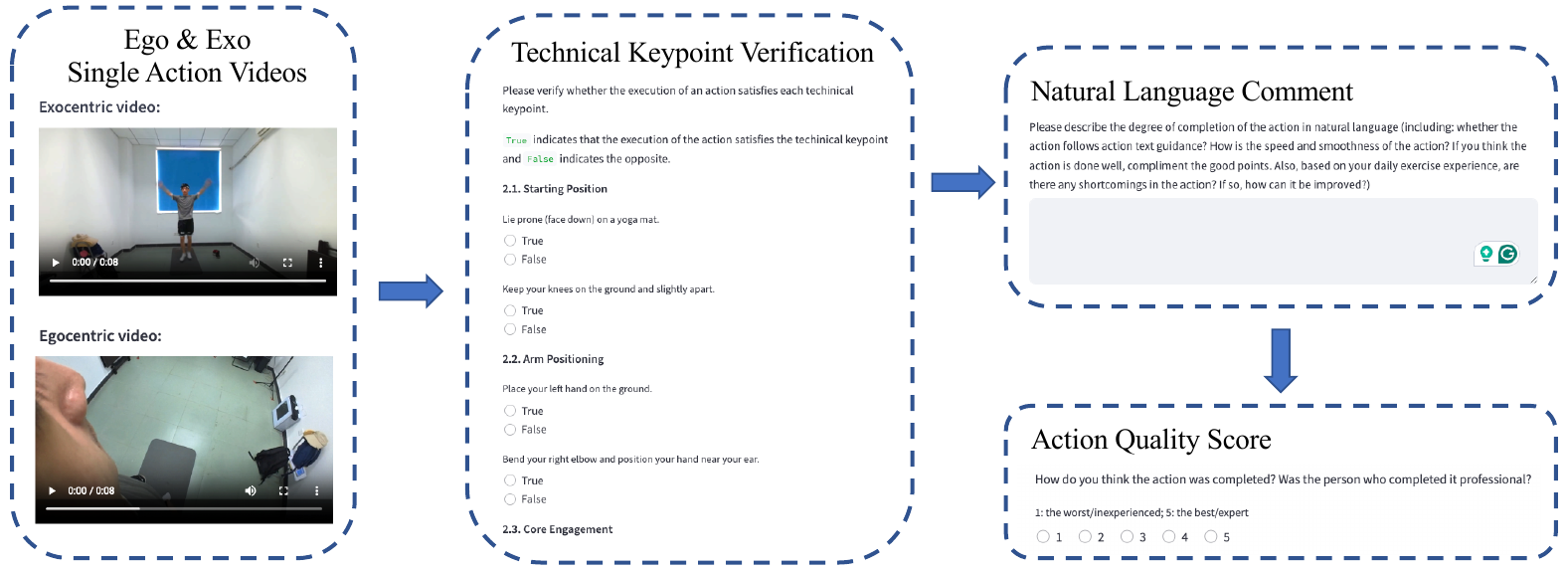}
  \caption{The annotation process of interpretable action judgement. }
  \label{fig:annotation_process}
\end{figure}

\vspace{0.1cm}
\noindent\textbf{Annotation tools.}
We use the popular COIN \cite{coin} annotation tool for two-level temporal boundaries.
Besides, we develop a web-based annotation tool to collect the annotations of interpretable action judgment.
\cref{fig:annotation_process} introduce the workflow of the annotation process of interpretable action judgment. 

\vspace{0.1cm}
\noindent\textbf{Guidance and Technical Keypoints.}
As discussed in the main body of the paper, we obtain several technical keypoints from the text guidance. 
We use the text guidance provided in FLAG3D \cite{flag3d}. 
We use the prompt ``\emph{In this task, you are given text guidance of a fitness action. Your job is to separate the text guidance into several key points.}'' to require LLM (\ie, GPT-4) to extract technical keypoints from the text guidance.
\cref{tab:action_keypoints} shows an example of the extracted technical keypoints.

\begin{figure}[ht]
  \centering
  \includegraphics[width=1\linewidth]{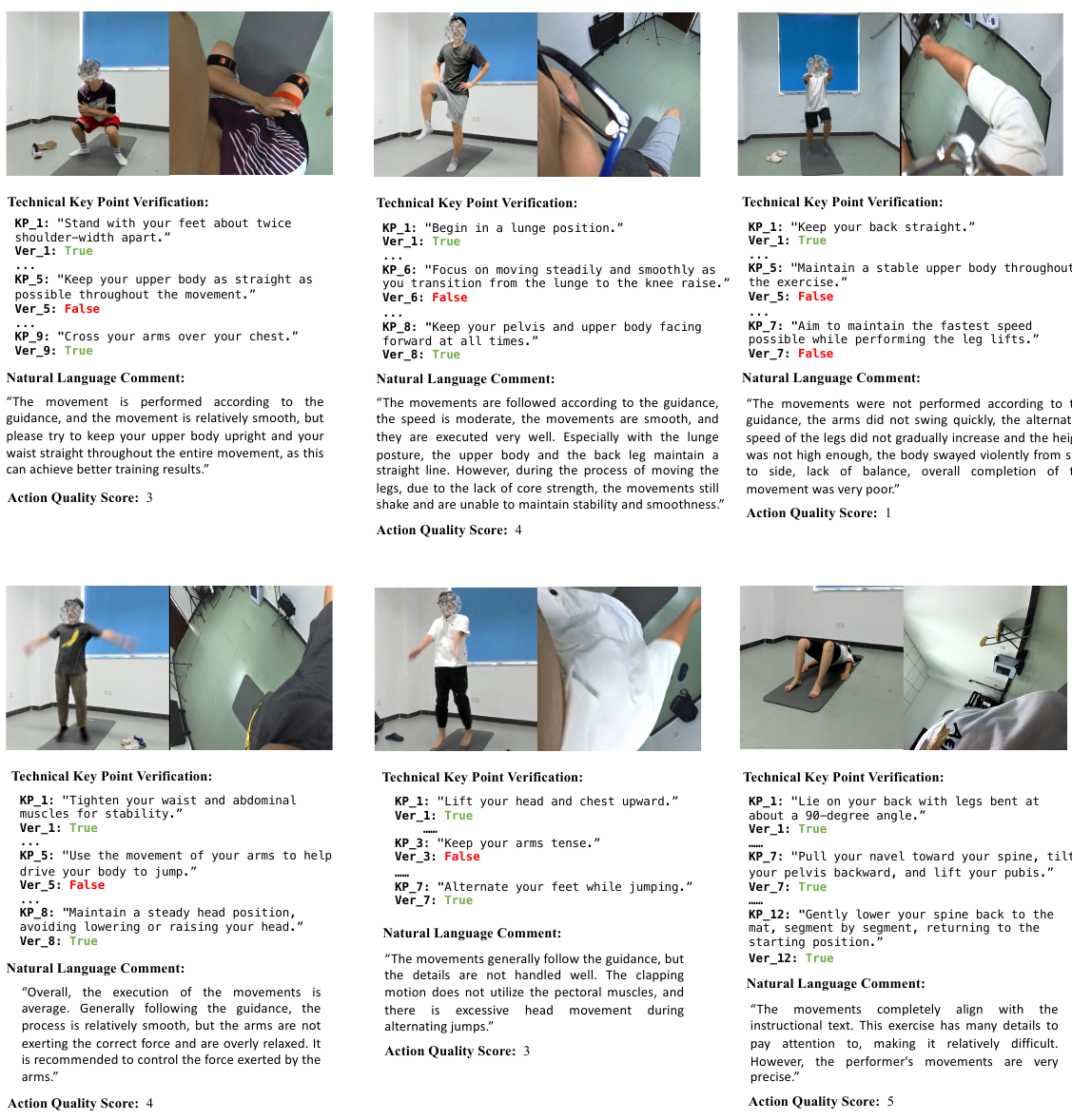}
  \caption{More examples of interpretable action judgement. }
  \label{fig:annotation_examples}
\end{figure}

\vspace{0.1cm}
\noindent\textbf{More Examples.}
We show more examples of annotations of interpretable action judgment in \cref{fig:annotation_examples}.

\vspace{0.1cm}
\noindent{\textbf{Privacy and Ethics.}} From the onset, privacy and ethics standards were critical to the data collection and release effort. All videos are recorded after we obtain the consent provided by participants. All human experts are asked to sign a privacy protection agreement to prevent data and privacy disclosure during the annotation process. To further protect the privacy and personal information, before the data release, we will ensure that the release resources do not contain privacy-sensitive content (\eg, real name). 

\begin{table}[t]
  \begin{center}
  \caption{\textbf{Recorded fitness actions.} Abbr.: the abbreviation of the fitness action.} 
  \label{tab:motion_category_appendix}
  \vspace{-0.2cm}
  \scalebox{0.7}{
  \begin{tabular}{lr||lr||lr}
  \toprule
  \textbf{Action types}                                & \textbf{Abbr.}  & \textbf{Action types}                                & \textbf
{Abbr.} & \textbf{Action types}                                & \textbf{Abbr.}\\
  \midrule
  \,\,1:\,Kneeling Push-ups                              & KPU            & \,\,5:\,Shoulder Bridge                                & SB & 
\,\,10:\,Jumping Jacks                                & JJ\\
  \,\,2:\,Push-ups                                       & PU             & \,\,6:\,Sit-ups                                        & SU & 
\,\,11:\,High Knee                                    & HK\\
  \,\,3:\,Kneeling Torso Twist                           & KTT            & \,\,7:\,Leg Reverse Lunge                             & LRL & 
\,\,12:\,Clap Jacks                                   & CJ\\
  \,\,4:\,Knee Raise and Abdomi-\,\,\,\,\,\,\,\,\,\,     & KRAMC          & \,\,8:\,Leg Lunge with Knee Lift\,\,\,\,\,\,\,\,\,\,  & LLKL\\
  \,\,\,\,\,\,\,nal Muscles Contract                     &                & \,\,9:\,Sumo Squat                                    & SS\\
  \bottomrule
  \end{tabular}
  }
  \end{center}
  \vspace{-0.3cm}
\end{table}

\begin{table}[t]
    \centering
    
    \caption{\textbf{All recorded fitness sequences.} For the correspondence between action names and abbreviations, please refer to \cref{tab:motion_category_appendix}. SID: sequence ID.}
    \vspace{-0.2cm}
    \label{tab:sequence_list}
    \scriptsize
    \resizebox{1\linewidth}{!}{
    \begin{tabular}{ll|ll|ll} 
    \toprule
    \textbf{SIDs} & \textbf{Action Orders} & \textbf{SIDs} & \textbf{Action Orders} & \textbf{SIDs} & \textbf{Action Orders}\\
    \midrule
    \,\,1 & KPU, SU, JJ                     &  \,\,30 & KTT, KPU, KRAMC, SU, JJ, HK      & \,\,59 & SB, CJ, KRAMC, HK, SS \\ 
    \,\,2 & KPU, JJ, SU                     &  \,\,31 & KPU, KTT, KRAMC, SU, JJ, HK  & \,\,60 & SB, CJ, KRAMC, KTT, SS \\ 
    \,\,3 & KTT, SU, JJ                     &  \,\,32 & KPU, KTT, KRAMC, HK, JJ, SU  & \,\,61 & SB, CJ, KRAMC, LLKL, SS \\
    \,\,4 & KTT, JJ, SU                     &  \,\,33 & KTT, KRAMC, SU, LLKL, JJ, HK & \,\,62 & SB, CJ, KRAMC, SU, SS \\ 
    \,\,5 & SU, KTT, JJ                     &  \,\,34 & KTT, JJ, SU, LLKL, KRAMC, HK         & \,\,63 & JJ, SB, KRAMC, CJ, SS \\ 
    \,\,6 & KPU, JJ, HK                     &  \,\,35 & KRAMC, KTT, SU, LLKL, JJ, HK         & \,\,64 & LLKL, SB, KRAMC, CJ, SS \\
    \,\,7 & KPU, HK, JJ                     &  \,\,36 & PU, LRL, CJ                          & \,\,65 & PU, SB, LRL, KRAMC, CJ, SS \\
    \,\,8 & JJ, KPU, HK                     &  \,\,37 & LRL, PU, CJ                          & \,\,66 & PU, SB, LRL, CJ, SS, KRAMC \\
    \,\,9 & KPU, SU, LLKL, JJ               &  \,\,38 & SB, CJ, LRL                          & \,\,67 & PU, LRL, SB, KRAMC, CJ, SS \\
    \,\,10 & KPU, SU, JJ, LLKL              &  \,\,39 & LRL, SB, CJ                          & \,\,68 & SB, PU, SU, LRL, CJ, SS \\
    \,\,11 & SU, KPU, LLKL, JJ              &  \,\,40 & PU, LRL, KRAMC, CJ                   & \,\,69 & SB, PU, LLKL, LRL, CJ, SS \\
    \,\,12 & KTT, SU, LLKL, JJ              &  \,\,41 & PU, LRL, CJ, KRAMC                   & \,\,70 & PU, SB, SU, SS, CJ, LRL \\
    \,\,13 & KTT, SU, JJ, LLKL              &  \,\,42 & LRL, PU, KRAMC, CJ                   & \,\,71 & PU, SB, JJ, SS, CJ, LRL \\
    \,\,14 & LLKL, SU, KTT, JJ              &  \,\,43 & SB, LRL, KRAMC, CJ           & \,\,72 & PU, SB, LLKL, SS, CJ, LRL \\
    \,\,15 & KPU, KTT, JJ, HK               &  \,\,44 & SB, LRL, CJ, KRAMC \         & \,\,73 & PU, SB, KTT, SS, CJ, LRL \\
    \,\,16 & KPU, KTT, HK, JJ               &  \,\,45 & KRAMC, LRL, SB, CJ           & \,\,74 & SB, HK, LRL, KRAMC, CJ, SS \\
    \,\,17 & KTT, KPU, JJ, HK               &  \,\,46 & PU, SB, CJ, SS               & \,\,75 & SB, CJ, LRL, KRAMC, JJ, SS \\
    \,\,18 & KPU, KTT, SU, LLKL, JJ         &  \,\,47 & PU, SB, SS, CJ               & \,\,76 & SU, SB, LRL, KRAMC, CJ, SS\\
    \,\,19 & KPU, KTT, SU, JJ, LLKL         &  \,\,48 & SB, PU, CJ, SS               & \,\,77 & LLKL,KRAMC,HK \\
    \,\,20 & KPU, SU, KTT, LLKL, JJ         &  \,\,49 & PU, SB, LRL, KRAMC, CJ       & \,\,78 & SB,LRL,CJ \\
    \,\,21 & KTT, KPU, SU, JJ, HK           &  \,\,50 & PU, SB, LRL, CJ, KRAMC       & \,\,79 & PU,SB,CJ,LRL \\
    \,\,22 & KPU, KTT, SU, JJ, HK           &  \,\,51 & PU, LRL, SB, KRAMC, CJ       & \,\,80 & KTT,SU,JJ,HK \\
    \,\,23 & KPU, KTT, HK, JJ, SU           &  \,\,52 & SB, PU, LRL, CJ, SS          & \,\,81 & KPU,KTT,SU,JJ \\
    \,\,24 & KTT, KRAMC, LLKL, JJ, HK       &  \,\,53 & PU, SB, LRL, CJ, SS          & \,\,82 & KPU,KTT,LLKL,SU,JJ,HK \\
    \,\,25 & KTT, JJ, LLKL, KRAMC, HK       &  \,\,54 & PU, SB, SS, CJ, LRL          & \,\,83 & KPU,KTT,JJ,HK,LLKL \\
    \,\,26 & KRAMC, KTT, LLKL, JJ, HK       &  \,\,55 & SB, HK, KRAMC, CJ, SS        & \,\,84 & PU,SB,KRAMC,CJ,SS \\
    \,\,27 & KPU, KTT, SU, LLKL, JJ, HK     &  \,\,56 & SB, KTT, KRAMC, CJ, SS       & \,\,85 & KPU,SU,KTT,KRAMC,JJ \\
    \,\,28 & KPU, KTT, SU, JJ, HK, LLKL     &  \,\,57 & SB, KPU, KRAMC, CJ, SS       & \,\,86 & KTT,KRAMC,LLKL,SU,JJ,HK \\
    \,\,29 & KPU, SU, KTT, LLKL, JJ, HK     &  \,\,58 & SB, CJ, KRAMC, KPU, SS \\
    \bottomrule
    \end{tabular}
    }
\end{table}

\begin{table}[t]
    \centering
    \caption{Examples of the language guidance and technical keypoints.}
    \label{tab:action_keypoints}
    \small
    \setlength{\extrarowheight}{2pt}%
    \resizebox{1\linewidth}{!}{
    \begin{tabular}{p{2cm}p{9.5cm}}
    \toprule
    \textbf{Action Name}        & Clap-Jacks                         \\
    \cmidrule{2-2}
    \textbf{Language Guidance}  & {Lift your head and chest, and tense your abdomen. tense the arms, and use the strength of your pectoral muscles to clap your hands while jumping back and forth with alternating feet.
    }\\  
    \cmidrule{2-2}
    \textbf{Technical Key Points }        &      
    \text{\,\,\,\,\,\,}\textbf{KP\_1:} Lift your head and chest upward.\newline
    \text{\,\,\,\,\,\,}\textbf{KP\_2:} Tense your abdominal muscles for stability.\newline
    \text{\,\,\,\,\,\,}\textbf{KP\_3:} Keep your arms tense.\newline
    \text{\,\,\,\,\,\,}\textbf{KP\_4:} Use the strength of your pectoral (chest) muscles.\newline
    \text{\,\,\,\,\,\,}\textbf{KP\_5:} Clap your hands while performing the exercise.\newline
    \text{\,\,\,\,\,\,}\textbf{KP\_6:} Perform jumping movements back and forth.\newline
    \text{\,\,\,\,\,\,}\textbf{KP\_7:} Alternate your feet while jumping.
    \\
    \midrule
    \textbf{Action Name}        & Sumo Squat                         \\
    \cmidrule{2-2}
    \textbf{Language Guidance}  & {Stand about twice shoulder-width apart, with your toes facing diagonally forward. when squatting to the thighs parallel to the ground, keep your knees in the same direction as your toes. keep your upper body as straight as possible, and sit back slightly when squatting. cross your arms over your chest.
    }\\  
    \cmidrule{2-2}
    \textbf{Technical Key Points }        &     
    \text{\,\,\,\,\,\,}\textbf{KP\_1:} Stand with your feet about twice shoulder-width apart.\newline
    \text{\,\,\,\,\,\,}\textbf{KP\_2:} Position your toes so they are pointing diagonally forward.\newline
    \text{\,\,\,\,\,\,}\textbf{KP\_3:} Squat down until your thighs are parallel to the ground.\newline
    \text{\,\,\,\,\,\,}\textbf{KP\_4:} Ensure your knees are aligned in the same direction as your toes.\newline
    \text{\,\,\,\,\,\,}\textbf{KP\_5:} Keep your upper body as straight as possible throughout the movement.\newline
    \text{\,\,\,\,\,\,}\textbf{KP\_6:} Slightly sit back as you squat down, like sitting into a chair.\newline
    \text{\,\,\,\,\,\,}\textbf{KP\_7:} Cross your arms over your chest.
    \\
    \bottomrule
    \end{tabular}
    }
\end{table}

\clearpage

\section{Benchmarks}
\label{sec:more_benchmark}
In this section, we will first present more details and experiments on Action Classification, Cross-View Sequence Verification, and Guidance-based Execution Verification. Then, we will introduce two more benchmarks on Action Localization and Cross-View Determination.

\subsection{Action Classification}
\label{sec:more_details_action_recognition}
\noindent{\textbf{Implementation.} 
(1) \textbf{Data Construction}: We select 4,753 single action videos (3,000 for training and 1,753 for testing) to construct the Action Classification benchmark.
(2) \textbf{Pre-trained weights}: We evaluate models with various pre-training strategies to construct action classification benchmark. For I3D \cite{i3d}, EgoVLP \cite{egovlp}, and TimeSformer \cite{timesformer} pre-trained on the K600 \cite{k600} dataset, we use the official pre-trained weights. For TimeSformer pre-trained on Ego-Exo4D \cite{egoexo4d}, we follow the setting of ``Key-Step Recognition'' benchmark in Ego-Exo4D to initialize the model with K600 pre-trained weights then trained on Ego-Exo4D.
(3) \textbf{Experiment Settings}: The input size of the video clip is set as $16\times 224\times 224$. During training, the video clips are sampled with temporal augmentation followed by random cropping. We train the models for 200 epochs with a base learning rate of 1e-5 and adopt a multi-step learning rate decay with a decay rate of 0.5 for every 25 epochs. For evaluation, a single video is uniformly sampled from the video, followed by center cropping.
}

\begin{table}[ht]
  \caption{\textbf{Action classification results on different views}. We report Top-1 accuracies on different veiws for TimeSformer model with Ego-Exo4D pre-training. }
  \label{tab:action_recognition_egoexo}
  \vspace{-0.3cm}
  \centering
  \resizebox{0.8\linewidth}{!}{
  \begin{tabular}{c|cccc|cccc}
      \toprule  
      \textbf{Train on} & \textbf{Exo-L} & \textbf{Exo-M} & \textbf{Exo-R} & \textbf{Exos} & \textbf{Ego-L} & \textbf{Ego-R} & \textbf{Ego-M} & \textbf{Egos}\\
      \midrule
      Exo & 0.8746 & 0.8993 & 0.8746 & 0.8825 & 0.0814 & 0.1017 & 0.0610 &  0.0814 \\
      Ego & 0.1559 & 0.1475 & 0.1763 & 0.1601 & 0.8305 & 0.8508 & 0.7186 & 0.8000\\
      Ego \& Exo  & 0.9051 & 0.8921 & 0.8949 &  0.8975 & 0.8475 & 0.7898 & 0.7153 & 0.7840 \\
  \bottomrule  
  \end{tabular}
}
\end{table}

\noindent{\textbf{More Experiment Analysis.}}
In the main paper's experiments, we found that models perform worse on egocentric data. In this section, we will explain these results more fully.
The first reason leading to this result is the invisibility of the human body. To support this view, we evaluate the performance of each view. As shown in \cref{tab:action_recognition_egoexo}, it is more difficult for a model to recognize an action from videos shot from the Ego-M camera (\ie, the forward-recording camera) than from other egocentric cameras (\ie, Ego-L and Ego-R). The main difference between videos shot from Ego-M and other egocentric Ego-cameras is that the human body is always out of view in videos from Ego-M.

Compared with Ego-M, videos shot from Ego-L and Ego-R record parts of the body. However, from \cref{tab:action_recognition_egoexo}, it can be observed that that the model still achieves poorer performance on videos shot from Ego-L and Ego-R than on those from exocentric cameras. To go deeper to this observation, we conduct a confusion evaluation. 
Specifically, we select one action (\ie, Leg Reverse Lunge) and two other actions (\ie, Knee Raise and Abdominal Muscles Contract, and Kneeling Torso Twist) whose egocentric videos are much easier to confuse models. The confusion matrixes and cropped frames are shown in \cref{{fig:why_ego_worse}}. From the egocentric video frames, similar action patterns (\ie, legs bending) can be observed among videos of these three actions, which cause serious confusion. On the contrary, the exocentric videos of these three actions are much more discriminating, which leads to higher classification performance. From these results, we conclude that the other reason leading to poorer full-body action understanding performance on egocentric videos is that it is easier to observe similar action patterns from egocentric videos, which will confuse models.

\begin{figure}[t]
  \centering
  \includegraphics[width=0.55\linewidth]{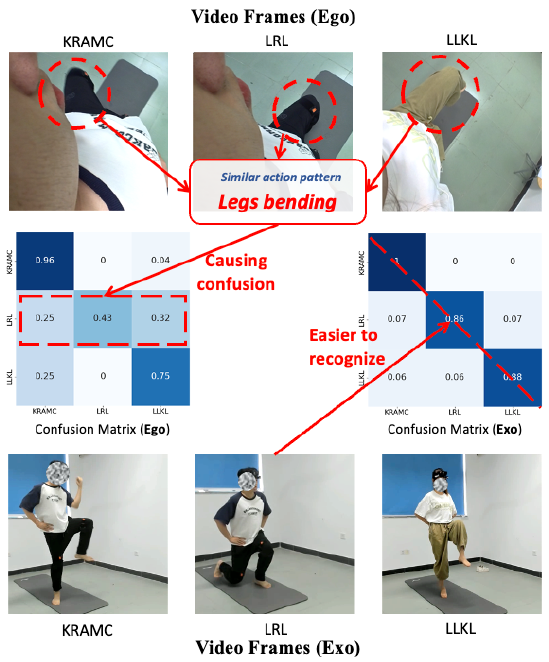}
  \caption{The similar action patterns observed by egocentric videos will confuse models to recognize an action. Best viewed in color. }
  \label{fig:why_ego_worse}
\end{figure}

\subsection{Cross-View Sequence Verification}
\label{sec:more_detail_cvsv}
\noindent{\textbf{More Details on Task Setup.}}
Following the task setup of existing work on SV \cite{svip,weakly-svr}, we formulate CVSV as a classification task during training, \ie, predicting the sequence class. 
During testing, the embedding distance $d$ (or similarity) between two videos indicates the verification score of this pair. 

Specifically, in training phase, a training set $D_{train}=\{(v_i,s_i)\}^N_{i=1}$ is used to construct a sequence classification task, where $v_i$ is a action sequence video and $s_i$ is a sequence label (\eg, a SID in \cref{tab:sequence_list}). 
Given a video $v\in \mathbb{R}^{3\times H\times W\times T}$ and its corresponding sequence label $s$, the model $f\odot g:{R}^{3\times H\times W\times T}\rightarrow \mathbb{R}^C$ is asked to predict the sequence label from $C$ sequence classes. Here $f$ is the embedding encoder, $g$ is the classifier. $H$, $W$, and $T$ are height, width, and the number of frames, respectively. 
In the testing phase, the model is asked to perform sequence verification on the test set where the sequence labels do not overlap with videos in the training set. Given a video pair $(v_i, v_j)$, a distance (or similarity) function $D$ is conducted on the embeddings of each video in the pair, which is denoted as $d_{ij} = D(f(v_i), f(v_j))$. A higher $d_{ij}$ indicates a lower possibility for $v_i$ to contain the same action sequence as $v_j$ (opposite if similarity function is used). In practical application, a threshold $\tau$ can be set to decide whether two sequences are consistent: if $d_{ij} > \tau $, sequences of $v_i$ and $v_j$ are consistent, otherwise inconsistent (opposite if similarity function is used).

\begin{figure}[t]
  \centering
  \includegraphics[width=0.8\linewidth]{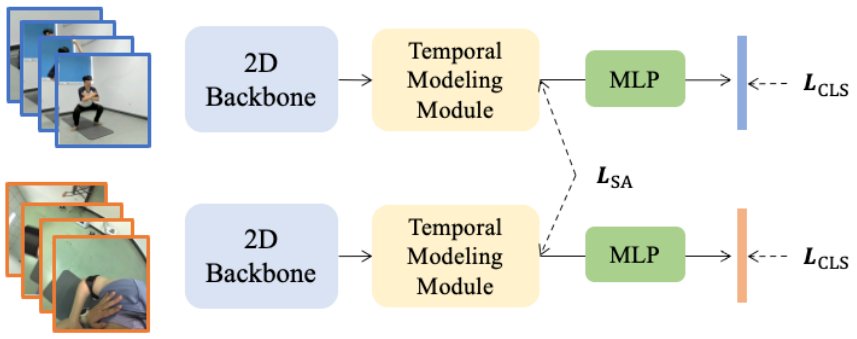}
  \caption{An overveiw of the CAT baseline for cross-view sequence verification. }
  \label{fig:cat_architecture}
\end{figure}

\vspace{0.3cm}
\noindent{\textbf{More Details about baseline model.}}
As discussed in the main paper, we adopt the state-of-the-art sequence verification model CAT \cite{svip} as the baseline model. The overview of CAT is shown in \cref{fig:cat_architecture}. The embedding encoder $f$ includes a 2D Backbone and a Temporal Modeling Module to encode video embeddings. The classifier $g$ is implemented as a Multi-Layer Perceptron. Specifically, the 2D backbone is implemented as a CLIP-ViT/16 \cite{clip}, and the Transformer encoder is adopted as a Temporal Modeling Module.
During training, CAT takes a pair of videos with the same action sequence as input and is optimized to learn to classify the action sequence labels by a classification loss $L_{CLS}$. 
Besides, an extra sequence alignment loss $L_{SA}$ is adopted to align video representations of videos with the same action sequence. 

\noindent{\textbf{Implementation.}} (1) \textbf{Data Construction}. Following previous works \cite{svip, weakly-svr,he2023collaborative}, we take 1074 action-sequence videos to build the CVSV dataset and make sure that the type of action sequences in the training set has no overlap with the test set. After that, we select 3,800 video pairs to train CAT and select another 3800 video pairs for testing.
(2) \textbf{Experiment Settings.} We follow the official setting of existing SV works \cite{svip, weakly-svr,he2023collaborative} to use the normalized Euclidean distance is used as the distance function. All experiments are conducted with a batch size of 8, a cosine learning rate scheduler with a base learning rate of 5e-5, and the models are trained for 40 epochs.

\subsection{Guidance-based Execution Verification}
\label{sec:more_details_gev}
\noindent{\textbf{More Details about GEVFormer.}}
This section will provide more details on implementing GEVFormer, including the architectures and loss formulation.

In GEVFormer, the TCM module is implemented as a 2-layer Transformer Encoder with 2-head attention. CMV module is designed as a 2-layer Transformer Decoder with 2-head attention and a linear evaluator. The prediction results $P=\{p_1,...,p_n\}$ is normalized by $Sigmoid(\cdot)$ function.

As discussed in the main paper, two losses are adopted to train GEVFormer (\ie, $L_{GEV}$ and $L_{Align}$).
First, given the predicted results $P$, the ground-truth targets are denoted as $P^{gt}=\{p^{gt}_1,...,p^{gt}_n\}$, where $p^{gt}_i$ is a binary value and $p^{gt}_i=1$ indicates that the execution of the action satisfies the i-th technical keypoint. After that, $L_{GEV}$ is implemented as a Binary Cross-Entropy loss:
\begin{equation}
  L_{GEV} = -\sum_{i=1}[p^{gt}_i logp_i + (1-p^{gt}_i)log(1-p_i)].
  \label{eq:L_gev}
\end{equation}
Besides, given a mini-batch of training samples $V=\{v_1,v_2,...,v_K\}$ ($K$ is the batch size), we randomly sample another batch of video $\widetilde{V} = \{v_1,...v_K\}$, where $v_i$ and $v_j$ are time-aligned (\ie, synchronized). After that, we fed videos in $V$ and $\widetilde{V}$ into GEVForrmer and get the enhanced visual embeddings (outputs of TCM module), which are denoted as $G=\{g_1,...,g_K\}$ and $\widetilde{G}=\{\widetilde{g}_1,...,\widetilde{g}_K\}$, respectively.
Given $G$ and $\widetilde{G}$, the synchronized video alignment loss $L_{Align}$ is written as:
\begin{equation}
  L_{Align} = \frac{1}{K}\sum^K_{i=1}log\frac{exp(\psi(g_i,\widetilde{g}_i) / \delta )}{\sum^K_{j=1}exp(\psi(g_i,\widetilde{g}_j) / \delta)},
  \label{eq:L_align}
\end{equation}
\begin{equation}
  \psi(g_i,g_j) = \frac{g_i}{||g_i||} \cdot \frac{g_j}{||g_j||}, 
\end{equation}
where $\psi(.,.)$ indicates cosine similarity function, and $\delta$ is the tempreture parameter.

\vspace{0.2cm}
\noindent{\textbf{More experiment settings.}} We select 3,260 samples from videos shot by Ego-R, Ego-L, Exo-R and Exo-L. After that, we split them into training set and test set (2,232 videos for training and 1,028 for testing). We use video frames sampled with a sample rate 1/16 as the input. During training, a random temporal augmentation is used to augment data. By default, $\lambda$ is set as 0.7.

\begin{table}[t]
  \caption{Ablation study on different components of GEVFormer.}
  \vspace{-0.3cm}
  \label{tab:ablation_components}
  \scriptsize
  \centering
  \scriptsize
  \resizebox{\linewidth}{!}{
  \begin{tabular}{c|ccc|ccc|ccc|c}
      \toprule
      \multirow{2}{*}{{Methods}} & \multirow{2}{*}{{TCM}} & \multirow{2}{*}{{CMV}} & \multirow{2}{*}{${L_{Align}}$} & \multicolumn{3}{c|}{{Exo}}       & \multicolumn{3}{c|}{{Ego}}       & {Avg}      \\
         &                  &                      &                              & F1-score & Precision & Recall & F1-score & Precision & Recall & F1-score \\ 
       \midrule
       CLIP-GEV &&   &   & 0.5080   & \textcolor{lightgray}{0.5362}    & \textcolor{lightgray}{0.4657} & 0.4780   & \textcolor{lightgray}{0.5401}    & \textcolor{lightgray}{0.4094} & 0.4881   \\ 
      &\checkmark &  &   &     0.5174     &  \textcolor{lightgray}{0.5407}   & \textcolor{lightgray}{0.4831}   &     0.5103     &    \textcolor{lightgray}{0.5492}       &    \textcolor{lightgray}{0.4890}    &      0.5138    \\
      &\checkmark & \checkmark  &   & 0.5282	& \textcolor{lightgray}{0.5502}	& \textcolor{lightgray}{0.5080}	& 0.5248	& \textcolor{lightgray}{0.5570}	& \textcolor{lightgray}{0.4960}	& 0.5265   \\ 
      GEVFormer &\checkmark & \checkmark & \checkmark & \textbf{0.5452}   & \textcolor{lightgray}{0.5219} & \textcolor{lightgray}{0.5707} & \textbf{0.5425} & \textcolor{lightgray}{0.5186} & \textcolor{lightgray}{0.5687} & \textbf{0.5439}  \\
      \bottomrule
  \end{tabular}
}
\end{table}

\vspace{0.2cm}
\noindent{\textbf{More Experiment Analysis.}}
In this section, we conduct ablation studies on GEVFormer. We start by ablating the components of GEVFormer. As shown in 
\cref{tab:ablation_components}, when adding each component from
the CLIP-GEV baseline to GEVFormer, performance gradually improved, showing each component's contribution.

\subsection{Action Localization}
\begin{table}[t]
    \centering
    \caption{{Temporal Action Locaization benchmark results on different views}. Results in \la{blue} and \ym{red} indicates the best performance on exocentric and egocentric videos, respectively.}
    \vspace{-0.3cm}
    \resizebox{0.75\linewidth}{!}{
    \begin{tabular}{c|c|ccccc|c}
    \toprule  
    {Train on} & {Test on} & {AP@0.3}  &  {AP@0.4} & {AP@0.5}  &  {AP@0.6} & {AP@0.7} & {mAP} \\
    \midrule
    \multirow{2}{*}{Ego} & Ego & 43.30 & 41.48 & 37.61 & 28.99 & 16.92 & 33.66\\
    \multirow{2}{*}{}  & Exo &  5.32 &4.62 &3.24& 1.89 &0.67 & 3.15\\
    \midrule
    \multirow{2}{*}{Exo}  & Ego & 4.79& 4.05 &3.01 &1.52 &0.61 & 2.80\\
    \multirow{2}{*}{} &  Exo & \la{49.87}  &\la{48.48} &\la{44.78} &\la{37.81} & {23.64} & \la{40.92}\\
    \midrule
    \multirow{2}{*}{Ego \& Exo} &  Ego &\ym{45.45} &\ym{43.21} &\ym{39.29} &\ym{32.15} &\ym{18.04} & \ym{35.63}\\
    \multirow{2}{*}{} &  Exo & 48.47 &46.82 &43.70 &36.22 & \la{23.65} & 39.77\\
    \bottomrule  
    \end{tabular}
    }
    \label{tab:tal_e50}
\end{table}

\noindent{\textbf{Task Setups.}}
TAL \cite{actionformer,tal_survey,tadtr} aims to identify action instances (\ie, foreground) in time and recognizing their categories.
Note that the most discriminating part of Fitness action is the ``executing'' stage.
Hence, in the Action Localization benchmark, we regard an action's “executing” step as the foreground,
otherwise as the background. The model is asked to predict all temporal boundaries and the action type of the foreground given an untrimmed action sequence video containing various actions.

\vspace{0.1cm}
\noindent{\textbf{Implementation.}}
\textbf{(1) Data Construction.} We select 1,165 untrimmed action sequence videos and randomly separate them into training and testing sets (66.7\% for training and 33.3\% for testing). 
\textbf{(2) Baseline Model.} We apply competing state-of-the-art transformer-based TAL method, TadTR \cite{tadtr}, using frame-wise features etxtracted from CLIP \cite{clip}. 
\textbf{(3) Matrics.} Performance is evaluated by mean average percision (mAP) at different intersections over union (IoU) thresholds of \{0.3, 0.4, 0.5, 0.6, 0.7\}.
\textbf{(4) Other experiment settings.} In our implementation, we use 10 action queries. Following previous work \cite{tadtr,bmn}, we crop each feature sequence with windows of length 450 and overlap of 75\%. We train TadTR on EgoExo-Fitness for 50 epochs with an inital learning rate of 1e-4.
For other experiment settings, we follow the official implementation of TadTR \cite{tadtr} on the THUMOS14 dataset \cite{thumos}.

\vspace{0.2cm}
\noindent{\textbf{Experiment.}}
The benchmark result on Action Localization is shown in \cref{tab:tal_e50}. In Action Localization, We have similar findings as in the Action Classification benchmark, such as jointly training the model on multi-view data will not benefit localization results on both egocentric and exocentric viewpoints (\ie, only performance on egocentric data achieves improvement). 

\subsection{Cross-View Skill Determination}
Given a pair of action videos, Skill determination \cite{raan,whos_better} aims at inferring which video displays more skill. Such a task has shown great potential for training humans and intelligent agents. 
Such a task will benefit the practical application of training humans and intelligent agents.
Although previous works have achieved significant progress, today's skill determination dataset is either collected from exocentric viewpoints (\eg, best) or egocentric(-like) viewpoints (\eg, epic-skill).
However, in practical application, the videos may come from various viewpoints, which poses a new challenge to skill determination. To address this issue, we extend the traditional skill determination to a cross-view manner (\ie, Cross-View Skill Determination).

\vspace{0.2cm}
\noindent{\textbf{Task Setups.}}
Following previous works \cite{whos_better,raan}, we formulate cross-view skill determination (CVSD) as a pair-wise ranking task.
In this setup, given a video pair $(v_i, v_j)$ where $v_i$ display more skill than $v_j$, 
our goal is to learn a ranking function $f(\cdot)$ such that $f(v_i) > f(v_j)$.

\vspace{0.2cm}
\noindent{\textbf{Implementation.}}
{\textbf{(1) Data Construction.}}
EgoExo-Fitness provides the action quality scores in annotations of interpretable action judgment. Based on this, we construct the Cross-view Skill-determination data using the following strategy.
First, we sample 3328 single action videos shot by Ego-R, Ego-L, Exo-R and Exo-L cameras and separate them into 1976 training videos and 1352 testing videos.
Second, for training videos, we construct video pairs by pairing videos with the same type of action. We do the same for testing videos.
Third, given a video pair $(v_i, v_j)$ and their corresponding action quality score $s_i$ and $s_j$, we regard it as a valid pair if $s_i > s_j + \theta$ is satisfied. Here $\theta$ is set as 1.5.
By following this strategy, we get 37680 valid pairs (25136 for training and 12544 for testing) for Cross-view Skill Determination.
{\textbf{(2) Baseline model.}}
We use the state-of-the-art skill determination model RAAN \cite{raan} as our baseline model.
\textbf{{(3) Experiment settings.}} Following previous works \cite{raan,whos_better}, we train an individual model for each task. We sample 500 frames from the videos using the image feature extracted by CLIP \cite{clip} as the input of RAAN. For those videos with less than 500 frames, we adopt zero paddings behind the CLIP features and carefully modify the attention module of RAAN to adapt to the masked input. 

\begin{table}[t]
    \caption{Benchmark results on Cross-View Skill Determination. ``ego/exo'' indicates independent models are trained on ego-only and exo-only data. ``ego+exo'' indicates that the model is trained on both egocentric and exocentric data.}
    \vspace{-0.3cm}
    \scriptsize
        \centering
        \scriptsize
        \resizebox{0.5\linewidth}{!}{
        \begin{tabular}{c|ccc}
        \toprule
        \multirow{2}{*}{{Methods}} & \multicolumn{3}{c}{{Acc}}\\ 
        \multirow{2}{*}{} & Ego-Ego & Exo-Exo & Ego-Exo \\
        \midrule
        Random         & 0.5000 & 0.5000 & 0.5000\\
        \midrule
        RAAN(ego/exo)  & \textbf{0.7386} & 0.7656 & -\\
        RAAN(ego+exo)  & 0.7072 & \textbf{0.7768} & \textbf{0.7241} \\
        \bottomrule
        \end{tabular}
        }
        
        \label{tab:skill_determination_clip}
\end{table}

\noindent{\textbf{Experiment.}}
The benchmark results of Cross-view Skill Determination are shown in \cref{tab:skill_determination_clip}.
We have similar findings as in Cross-view Sequence Verification benchmark, \ie, training models with all training pairs will not benefit performance on Ego-Ego pairs.

\section{More Comparisons with Related Datasets}
\label{sec:more_comparison_datasets}
\subsection{EgoExo-Fitness v.s. Ego4D}
For a fair comparison, in the main paper, we compare EgoExo-Fitness with a subset of Ego4D \cite{ego4d}, which contains scenarios of technical full-body actions. All selected scenarios are listed below: 
    \{\emph{Dancing, Working out at home, Basketball, Climbing, Outdoor technical climbing/belaying/rappelling (includes ropework), Swimming in a pool/ocean, Football, Going to the gym: exercise machine-class-weights, Yoga practice, Working out outside, Rowing, Skateboard/scooter, Baseball, Roller skating, Playing badminton, Table Tennis, Bowling}\}.

From \cref{tab:comprison_datasets} in the main paper, we find that the subset only contains a tiny fraction (about 172h) of videos in the whole Ego4D, which suggests that the egocentric full-body action understanding is rarely addressed even for the largest egocentric video datasets. Compared with Ego4D, EgoExo-Fitness contains synchronized ego-exo videos and novel annotations on how well a fitness action is performed (\ie, annotations of interpretable action judgment), which provides novel resources for future works on view characteristics, multi-view modeling, and action judgment for the egocentric vision community.

\begin{table}[t]
    \caption{{More comparison betweens the proposed EgoExo-Fitness and the concurrent Ego-Exo4D \cite{egoexo4d} dataset}. Compared with Ego-Exo4D, the proposed EgoExo-Fitness collects videos of a new scenario (\ie, fitness) and augments data with novel annotations of interpretable action judgment (\ie, text guidance and technical keypoint verification are not provided in Ego-Exo4D). }
    \vspace{-0.3cm}
    \label{tab:comprison_datasets_egoexo}
    \scriptsize
        \centering
        \scriptsize
        \resizebox{1\linewidth}{!}{
        \begin{tabular}{l|c|ccccc|c}
        \toprule
        \multirow{2}{*}{\textbf{Datasets}} & \multirow{2}{*}{\textbf{Scenarios}}     & \multirow{2}{*}{\textbf{Step}} & \textbf{Text} & \textbf{Keypoint} & \multirow{2}{*}{\textbf{Comment}} & \multirow{2}{*}{\textbf{Score}} &\multirow{2}{*}{\textbf{Duration}}\\ 

        \multirow{2}{*}{} & \multirow{2}{*}{}          & \multirow{2}{*}{} & \textbf{guidance} & \textbf{verification} & \multirow{2}{*}{} & \multirow{2}{*}{} & \multirow{2}{*}{}\\ 
        
        \midrule
        \multirow{8}{*}{Ego-Exo4D v1 \cite{egoexo4d}}   & Cooking & \checkmark & & & \checkmark & \checkmark & 654h \\
        \multirow{8}{*}{}           & Health  & \checkmark & & & \checkmark & \checkmark & 124h \\
        \multirow{8}{*}{}           & Bike Repair & \checkmark & & & \checkmark & \checkmark & 83h \\
        \multirow{8}{*}{}           & Music &  &  & & \checkmark & \checkmark & 216h \\
        \multirow{8}{*}{}           & Basketball &  & & & \checkmark & \checkmark & 61h \\
        \multirow{8}{*}{}           & Climbing &  &  & & \checkmark & \checkmark & 88h \\
        \multirow{8}{*}{}           & Soccer &  &  & & \checkmark & \checkmark & 96h\\
        \multirow{8}{*}{}           & Dancing &  &  & & \checkmark & \checkmark & 99h\\
        \midrule
        \multirow{8}{*}{Ego-Exo4D v2 \cite{egoexo4d}}   & Cooking & \checkmark & & & \checkmark & \checkmark & 564h \\
        \multirow{8}{*}{}           & Health  & \checkmark & & & \checkmark & \checkmark & 114h \\
        \multirow{8}{*}{}           & Bike Repair & \checkmark & & & \checkmark & \checkmark & 82h \\
        \multirow{8}{*}{}           & Music &  &  & & \checkmark & \checkmark & 180h \\
        \multirow{8}{*}{}           & Basketball &  & & & \checkmark & \checkmark & 78h \\
        \multirow{8}{*}{}           & Climbing &  &  & & \checkmark & \checkmark & 93h \\
        \multirow{8}{*}{}           & Soccer &  &  & & \checkmark & \checkmark & 66h\\
        \multirow{8}{*}{}           & Dancing &  &  & & \checkmark & \checkmark & 106h\\
        \midrule
        \cellcolor{yellow}EgoExo-Fitness(Ours)    & \cellcolor{yellow}Fitness    & \cellcolor{yellow}\checkmark & \cellcolor{yellow}\checkmark & \cellcolor{yellow}\checkmark & \cellcolor{yellow}\checkmark & \cellcolor{yellow}\checkmark & \cellcolor{yellow}32h\\
        \bottomrule
        \end{tabular}
        }
\end{table}

\subsection{EgoExo-Fitness v.s. Ego-Exo4D}
\label{sec:appendix_compare_egoexo}
As supplements to \cref{tab:comprison_datasets_egoexo_partial}, we provide more comparisons between our datasets and Ego-Exo4D \cite{egoexo4d} in \cref{tab:compare_egoexo4d}.
Besides, beyond the similarities and differences discussed in \cref{sec:comparison_datasets}, our dataset has a comparative scale with each scenario of \emph{full-body (physical)} actions in Ego-Exo4D (see the \cref{tab:compare_egoexo4d}). Note that for fair comparisons, single actions recorded by RGB cameras are considered.

We hope the proposed EgoExo-Fitness can be another resource for studying egocentric full-body action understanding and skill guiding.

\begin{table}[ht]
  \caption{Comparisons between Ego-Exo4D \cite{egoexo4d} on dataset scale. Our dataset has a comparative scale with each scenario of \emph{full-body (physical)} actions in Ego-Exo4D}
  \vspace{-0.3cm}
  \centering
  \resizebox{0.7\linewidth}{!}{
  \begin{tabular}{c|cccc|c}
  \toprule  
  Datasets & \multicolumn{4}{c|}{Ego-Exo4D v2 \cite{egoexo4d}} & \cellcolor{yellow}{Ours}\\
  Scenarios &  Basketball & Climbing & Soccer &  Dancing & \cellcolor{yellow}{Fitness}\\
  \midrule
  \# Tasks/Action Types & 3 & 11 & 3 & 2 & \cellcolor{yellow}12 \\
  \# Single Actions(RGB) & 4550 & 7191 & 1567 & 4367 & \cellcolor{yellow}6131\\
  \bottomrule  
  \end{tabular}
}
  \label{tab:compare_egoexo4d}
\end{table}

\begin{table}[ht]
  \caption{More comparison with existing full-body action datasets. EgoExo-Fitness has a comparative scale with existing related datasets.}
  \vspace{-0.3cm}
  \centering
  \resizebox{\linewidth}{!}{
  \begin{tabular}{c|cccccc|c}
  \toprule  
  Datasets &  MTL-AQA\cite{mtl-aqa} & FineGym\cite{finegym} & FineDiving\cite{finediving} & FLAG3D(real)\cite{flag3d} & 1st-basketball\cite{1st-basketball} & WEAR\cite{wear} & \cellcolor{yellow}{Ours}\\
  \midrule
  \# Videos & 1412 & 303 & 3000 & 7200 & 48 & 18 & \cellcolor{yellow}{1276}\\
  \# Single Actions & 1412 & 32697 & 3000 & 7200 & - & 615 & \cellcolor{yellow}6131\\
  \bottomrule  
  \end{tabular}
}
  \label{tab:compare_others}
\end{table}

\subsection{EgoExo-Fitness v.s. other related datasets}
We also provide more comparisons with existing datasets in \cref{tab:compare_others} as supplements to \cref{tab:comprison_datasets}, which show that our dataset has a comparative scale with existing full-body action datasets. 

\end{document}